\documentclass[preprint, 3p, times]{elsarticle}
\usepackage[utf8]{inputenc}

\usepackage{custom-preamble}
\usepackage{graphicx}
\usepackage{subcaption}
\usepackage{booktabs}
\usepackage{multirow}
\usepackage{pdflscape}
\usepackage{algorithmic}
\crefname{appendix}{}{}
\biboptions{sort,square}
\usepackage{xcolor}
\newcommand{\mytcp}[1]{%
  \textcolor{black}{\textsf{/*} #1 \textsf{*/}}\par
}
\usepackage{natbib}
\begin{document}

\begin{frontmatter}


\title{A reinforcement learning guided hybrid evolutionary algorithm for the latency location routing problem}

\author[inst1]{Yuji Zou}
\ead{yujizou6@gmail.com}

\author[inst1]{Jin-Kao Hao\corref{cor1}}
\ead{jin-kao.hao@univ-angers.fr}

\author[inst2]{Qinghua Wu}
\ead{qinghuawu1005@gmail.com}

\affiliation[inst1]{
    organization={LERIA, Université d'Angers},
    addressline={2 Boulevard Lavoisier}, 
    city={Angers},
    postcode={49045}, 
    country={France}
}
\affiliation[inst2]{
    organization={School of Management, Huazhong University of Science and Technology},
    addressline={No. 1037, Luoyu Road}, 
    city={Wuhan},
    postcode={},
    country={China}
}

\cortext[cor1]{Corresponding author}

\begin{abstract}

The latency location routing problem integrates the facility location problem and the multi-depot cumulative capacitated vehicle routing problem. This problem involves making simultaneous decisions about depot locations and vehicle routes to serve customers while aiming to minimize the sum of waiting (arriving) times for all customers. To address this computationally challenging problem, we propose a reinforcement learning guided hybrid evolutionary algorithm following the framework of the memetic algorithm. The proposed algorithm relies on a diversity-enhanced multi-parent edge assembly crossover to build promising offspring and a reinforcement learning guided variable neighborhood descent to determine the exploration order of multiple neighborhoods. Additionally, strategic oscillation is used to achieve a balanced exploration of both feasible and infeasible solutions. The competitiveness of the algorithm against state-of-the-art methods is demonstrated by experimental results on the three sets of 76 popular instances, including 51 improved best solutions (new upper bounds) for the 59 instances with unknown optima and equal best results for the remaining instances. We also conduct additional experiments to shed light on the key components of the algorithm. 

\emph{Keywords:} Routing; Latency location routing; Cumulative capacitated vehicle routing; Heuristics; Learning-driven optimization.
\end{abstract}

\end{frontmatter}

\section{Introduction}
\label{sec Introduction}

The location routing problem (LRP) plays a critical role in logistics management. The problem can be viewed as consisting of two sub-problems: the facility location problem FLP (i.e., selecting which depots to open) and the multi-depot vehicle routing problem (i.e., minimizing travel distance or other distance-related costs). The complexity of this problem arises from the need to consider both sub-problems simultaneously. The latency location routing problem (LLRP) is a LRP variant where the objective function of the underlying routing problem is to minimize the total waiting time of all customers. This customer-centric problem has many applications in different contexts related to emergency logistics operations in post-disaster relief, last-mile delivery with shared intermediate facilities in urban logistics, and delivery of perishable products \citep{ngueveu2010effective, nucamendi2022new}.

The LLRP can be thought of as a combination of the FLP and the multiple depot cumulative vehicle routing problem (MDCCVRP). Given that both constituent problems are NP-hard, the LLRP is inherently a computationally challenging problem \citep{moshref2016latency}. The problem can be defined on a complete graph $G=(V,E)$ with $V=D  \cup C$ and $E=\{(i,j): i,j \in V\}$, where $D$ is the set of homogeneous uncapacitated depots ($|D| \geq 1$) and $C=\{C_1,C_2,...C_m\}$ is the set of customers. Furthermore, $E$ is associated with a symmetric non-negative matrix $Y=(d_{ij})$ for the edges $(i,j)$, where $d_{ij}$ represents the travel time (or equivalently the distance) between two vertices, obeying the triangular inequality. There is a fleet $H$ of $N_v$ homogeneous vehicles, each with a capacity $P$. Each customer $i \in C$ has a demand $p_i$ that is fulfilled when a vehicle visits the customer. A solution to the LLRP problem, after determining the opening of at most $N_d$ depots, involves at most $N_v$ disjoint Hamiltonian tours from these depots.
Each tour starts and ends at the same opened depot, ensuring that each customer is visited exactly once by a tour. In addition, the sum of the demands of the customers in each tour must not exceed the capacity $P$ of the vehicle of the tour. Given a feasible solution, let $t^k_i$ be the arrival time of vehicle $k$ at customer $i$ ($t^k_i = 0$ if $i$ is not served by $k$). Then the LLRP is to find a solution $S$ that minimizes the sum of the waiting times of all customers.

\begin{center}
\begin{equation}
  \label{objective equation}
    Minimize \quad f(S) =\sum_{k \in H}{\sum_{i \in C}{t_i^k}}, \quad S \in \Omega
  \end{equation}
\end{center}

where $\Omega$ is the search space of all feasible candidate solutions for a LLRP instance. A mathematical formulation of the LLRP is provided in \citep{moshref2016latency}.

The LLRP was formally defined by Moshref-Javadi and Lee \cite{moshref2016latency}, where two algorithms were proposed to tackle it, a memetic algorithm (MA) and a recursive granular algorithm (RGA). MA uses a solution representation consisting of two parts: the first part represents the open depots, and the second part defines the vehicle assignment and the sequence of the customers visited by each vehicle. Four different initialization methods are applied to diversify the initial population. The Order Crossover (OX) is used to generate offspring solutions, and a local search based on three move operators is applied to improve the generated offspring. In RGA, a granularity-based neighborhood search method systematically modifies the current solution, followed by a local search procedure using three move operators to further improve the solution. The authors tested the algorithm on three benchmark sets for the LRP variants, specifying the number of open depots and assigned vehicles for each instance.

Nucamendi-Guillén \textit{et al.} \cite{nucamendi2022new} proposed two novel mixed-integer formulations based on multi-level networks for the LLRP. Additionally, they introduced a variant of the LLRP that considers the open cost of depots. To solve the LLRP, they presented a GRASP-based iterated local search (GBILS), which includes a constructive procedure and an improvement procedure. Within each iteration of the algorithm, a feasible solution is constructed by the constructive procedure and subsequently improved by the improvement procedure. The constructive procedure randomly applies different methods to determine the opened depots. Following this, customers are selected based on their distance to the opened depots, and a given number of routes are constructed. The remaining customers are assigned to routes based on both distance and remaining vehicle capacity. In the improvement procedure, three intra-route moves are iteratively applied until the solution can no longer be improved. Then, two inter-route moves are applied to further improve the solution. Experimental results showed that GBILS was able to find several new best-known solutions.

Osorio-Mora \textit{et al.}\cite{osorio2023effective} presented three algorithms that integrate simulated annealing (SA) and variable neighborhood descent (VND) \citep{mladenovic1997variable} to effectively solve the LLRP. SA serves as the method to escape local optima, and the VND procedure is applied to improve the solution. Upon reaching a threshold indicating that the solution cannot be further improved, the Lin-Kernighan-Helsgaun (LKH-3) heuristic \citep{helsgaun2017extension} is used to improve each individual route, addressing the corresponding CCVRP. The study introduced three types of VND methods exhibiting different behaviors. Experimental results on various instances indicated that the proposed algorithm significantly outperformed the state-of-art algorithms from previous studies in the domain.

Osorio-Mora \textit{et al.} \cite{osorio2023iterated} introduced an iterated local search (M-ILS) to tackle three latency vehicle routing problems with multiple depots, including the MDCCVRP, the LLRP and the multi-depot \textit{k}-traveling repairman problem. Recognizing that the proper selection of depots is critical to the success of the algorithm, the authors introduced a method that integrates LKH-3 and integer linear programming to simultaneously consider depot selection and vehicle routing. Following this, an iterative process applies a perturbation procedure, a local search based on five moves, and a SA-VND approach similar to \cite{osorio2023effective} to continually improve the quality of the solution. Then, the LKH-3 algorithm is used again to solve the CCVRP for each open depot. Experimental results showed that M-ILS stands out as the most powerful algorithm for solving the LLRP, consistently producing the best-known solutions for most benchmark instances. 

These reviewed studies have continuously advanced the state of the art in solving the LLRP. However, compared to other popular routing problems, research on the problem is still limited, and additional efforts are needed to develop more powerful and robust methods capable of finding satisfactory solutions for the most challenging problem instances. 

In this paper, we present a reinforcement learning guided hybrid evolutionary algorithm to address the LLRP, which includes a multi-parent edge assembly crossover and a learning-driven local search. Inspired by the edge assembly crossover for the traveling salesman problem \citep{nagata2013powerful}, the proposed crossover builds offspring solutions by inheriting subtours that contribute to high-quality solutions from the parents, preserving the desired solution diversity with multiple parents, and considering edge orientation during the crossover process. The VND-based local search is reinforced by two original techniques. It uses reinforcement learning to dynamically determine the exploration order of the underlying neighborhoods. It additionally adopts strategic oscillation \citep{glover2011case} to allow the VND procedure to achieve a balanced exploration between both feasible and infeasible search spaces. In terms of methodological contributions, the idea of the multi-parent edge assembly crossover can be conveniently applied to other routing problems, while the dynamic exploration of multiple neighborhoods with reinforcement learning is valuable for local search algorithms using multiple move operators.

In terms of computational contributions, we present experimental results on 76 popular benchmark instances to evaluate the performance of the algorithm. The results show that the algorithm is highly competitive with state-of-the-art algorithms, by finding 51 record-breaking results (new upper bounds) and matching all the remaining best-known results. These updated results are useful for future studies of the problem. Moreover, these results are achieved with shorter computation times than state-of-the-art methods, indicating its computational efficiency. We also perform experiments to understand the behavior of the algorithm. Finally, the codes of the algorithm will be made publicly available, which can be used by practitioners and researchers to solve related problems.

In the remainder of the paper, we present the proposed algorithm in Section \ref{sec rlhea}, a comprehensive computational comparison with leading algorithms in Section \ref{sec computation}. Section \ref{sec analyze} shows additional experiments to analyze the main algorithmic components and provide insights into their roles. Section \ref{sec conclusion} offers conclusions and outlines future work.

\section{Reinforcement learning guided hybrid evolutionary algorithm}
\label{sec rlhea}

The proposed reinforcement learning guided hybrid evolutionary algorithm (RLHEA) for the LLRP follows the framework of the population-based memetic algorithms (MAs) \citep{Moscato1999}, especially MAs in discrete optimization \citep{hao2012memetic}. MAs benefit from the synergy of these two complementary search strategies and provide a powerful framework for solving difficult problems. In particular, MAs have been very successful in solving several complex routing problems \citep{he2023general,he2023memetic,lu2018hybrid,nagata1997edge,ren2023effective}. RLHEA is an advanced MA characterized by its multi-parent edge assembly crossover (MPEAX) and its reinforcement learning guided variable neighborhood descent with strategic oscillation (RL-SOVND). It also includes a population initialization procedure, a mutation procedure, and a population management method. 

%.including the split delivery vehicle routing problem \citep{he2023general}, the multiple traveling salesman problem \citep{he2023memetic}, the traveling salesman problem with hotel selection \citep{lu2018hybrid}, and the multiple traveling repairman problem with profits \citep{ren2023effective}. the capacited vehicle routing problem \citep{nagata1997edge}

\subsection{Main scheme} 
\label{subsec scheme}

The general RLHEA framework is outlined in Algorithm \ref{Algo_main}. The learning functions $Q$ and $R$ are initialized at the beginning (line 2). The population $Pop$ is generated by the initialization procedure (line 3). After recording the best feasible solution $S_b$ found so far (line 4), the algorithm enters the "\textit{while}" loop to improve the population. In each generation, three parents are randomly selected from the population (line 6). The multi-parent edge assembly crossover is then employed to generate an offspring solution (line 7). If the offspring is infeasible (i.e., violating the vehicle capacity or/and the number of opened depots), it is immediately repaired (line 9), followed by a mutation procedure to diversify the solution (line 10). Then, RL-SOVND is activated to improve the quality of the offspring (line 11). The search information is updated based on the local optimum obtained (lines 12-15), and the population is updated accordingly (line 16). During the search process, if the best solution $S_b$ remains unchanged for a given number of consecutive generations, half of the individuals in the population are regenerated to introduce diversity (line 18). The algorithm terminates and returns the best feasible solution $S_b$ when reaching the predefined maximum number of generations (line 19).

\begin{algorithm}[ht]
    \footnotesize
    \caption{Pseudo-code of RLHEA}
    \label{Algo_main}
   % \begin{algorithmic}[1]
        \Input{ Problem instance, population size $\tau$, population replacement threshold $I_r$.}
        \Output{ The best solution $S_{b}$ found.}
        
        $I_n \gets 0$ \mytcp{Counter of consecutive generations the best solution $S_{b}$ is not improved}
         $Q,R \gets$ Initialize the Q-learning functions $Q$ and $R$ \;
         $Pop = \{S_1, S_2,...,S_{\tau}\} \leftarrow $ PopInitialize()  \mytcp{Population initiation, section \ref{subsec initialization}}%
         $S_b \gets {arg\ min}_{\forall{S_i} \in \text{Pop}} f(S_i)$ \mytcp{Record the best feasible solution found so far}

        \While{\textit{stopping condition is not reached}}{
             $S_A,S_B,S_C \gets $RandomParentSelection(P) \; 
             $S \gets $ MPEAX ($S_A,S_B,S_C$) \mytcp{Crossover, section \ref{subsec mpeax}}
            \If{\textit{Is\_Infeasible}($S$)}{
                 $S \gets$ Repair($S$) \mytcp{Repairing infeasibility, section \ref{subsec repair}}
                 }
            
             $S \gets$ Mutation($S$) \mytcp{Mutation, section \ref{subsec mutation}}
             $S_l,Q,R \gets$ RL-SOVND($S,Q,R$) \mytcp{Local improvement, section \ref{subsec rlsovnd}}
            \If{$f(S_l) < f(S_b)$}{
                 $S_b \gets S_l, I_n \gets 0 $ \;}
            \Else{
                 $I_n \gets I_n + 1$ \;}
             UpdatingPop($Pop,S_l$) \mytcp{Population management, section \ref{subsec population}}
            \If{$I_n > I_r$}{
                 ReplacingPop($Pop$), $I_n \gets 0$ \mytcp{Population replacement}}
            }
        \Return $S_{b}$ \mytcp{Return the best feasible solution found during the search}
        
   % \end{algorithmic}
\end{algorithm}

\subsection{Population initialization}
\label{subsec initialization}

The population initialization is a two-step process: depot selection and route construction, and applies a random initialization method and a greedy initialization method with an equal probability.  The first step randomly selects a predefined number of depots to open. For the route construction step, note that the objective value decreases as the number of vehicles increases since the edge returning from the last customer to the depot doesn't contribute to the objective. Additionally, the weight of an edge within a route affects the waiting time of all customers following that edge. Therefore, it is important that the edge at the beginning of each route is as short as possible, while utilizing all available vehicles and maintaining a balanced distribution of customers across all routes. Building upon these considerations, the second step assigns customers to different vehicles in a cyclic manner until all customers are served. Both greedy and random methods are used to assign customers. The greedy method first selects the shortest edge between the opened depots and the unselected customers to determine the depot and the first node until the specified number of routes is initialized. Then, each route is constructed by choosing the shortest edges between the last node of the route and the remaining unselected customers. The random method constructs the routes by selecting random depots and random customers, without using any greedy selection criterion. After a solution is constructed, it is improved using the RL-SOVND procedure, and then inserted into the population if no copy of the solution exists in the population. %When the number of solutions in the population reaches the predefined size $\tau$ (set to 20), the procedure terminates.

% However, there is a difference in the customer assignment process between the two approaches. The greedy approach first selects the shortest edge between the opened depots and the unselected customers to determine the depot and the first node until the specified number of routes is built. Then, each route is constructed by choosing the shortest edges between the last node of the route and the remaining unselected customers until all customers are assigned. In contrast, the random method constructs the routes by selecting random depots and random customers, without using any specific greedy selection criterion. After a solution is constructed, it is improved using the RL-SOVND procedure, and then inserted into the population unless the same solution already exists in the population. When the number of solutions in the population reaches the predefined size $\tau$ (set to 20), the procedure terminates.

\subsection{Offspring generation based on MPEAX}
\label{subsec mpeax}

%Crossover is a key component of memetic algorithms used to produce new solutions from existing ones. 

A meaningful crossover should be able to produce promising offspring by inheriting good features from the parents \citep{hao2012memetic}. Therefore, it is important to find good features that contribute to the high-quality of solutions and to pass them on to the offspring. For the traveling salesman problem and routing problems, the common edges shared by the parents are regarded as the key feature of high-quality solutions, and this feature has enabled the design of powerful crossover operators such as the maximal preservative crossover \citep{muhlenbein1991evolution}, the partition crossover \citep{whitley2009tunneling,sanches2017building} and the edge assembly crossover (EAX) \citep{nagata1997edge,nagata2013powerful}. Also, EAX-like operators also performed well on the capacitated vehicle routing \citep{nagata2007edge} and other well-known routing problems \citep{nagata2010penalty,he2023general,he2023memetic}. 

For the LLRP, we introduce the multi-parent edge assembly crossover (MPEAX) that relies on the idea of the original EAX crossover for the TSP \citep{nagata1997edge,nagata2013powerful}. MPEAX also generalizes the dEAX crossover of the two-individual evolutionary algorithm (TIEA) for the MDCCVRP \citep{zou2024two}. 

EAX for the TSP uses the joint graph (undirected graph) of the parent solutions to generate the so-called AB-\textit{cycles}, where an AB-\textit{cycle} is a cycle consisting of edges taken alternately from the parents and constitutes one core element of EAX. For the LLRP, recognizing that the direction of the route significantly impacts the objective function, we account for the route direction and use a directed graph to represent a LLRP solution. In this graph, each customer node is connected to one in-degree edge and one out-degree edge. Based on this graph representation of solutions, the proposed MPEAX crossover generalizes the notion of AB-\textit{cycle} to the case of directed edges with three parent solutions. In addition, the presence of  multiple depots in the LLRP may make it impossible to form an AB-\textit{cycle} due to the absence of edges with the same degree related to some depots. To overcome this, we treat all depots as a single node in our approach like in \citep{zou2024two} for the MDCCVRP. This treatment may not result in a strict 'cycle'. So we adopt the term AB-\textit{sequence} to accommodate this modification.

\begin{figure}
	\centering
	\includegraphics[width= 1.0\textwidth]{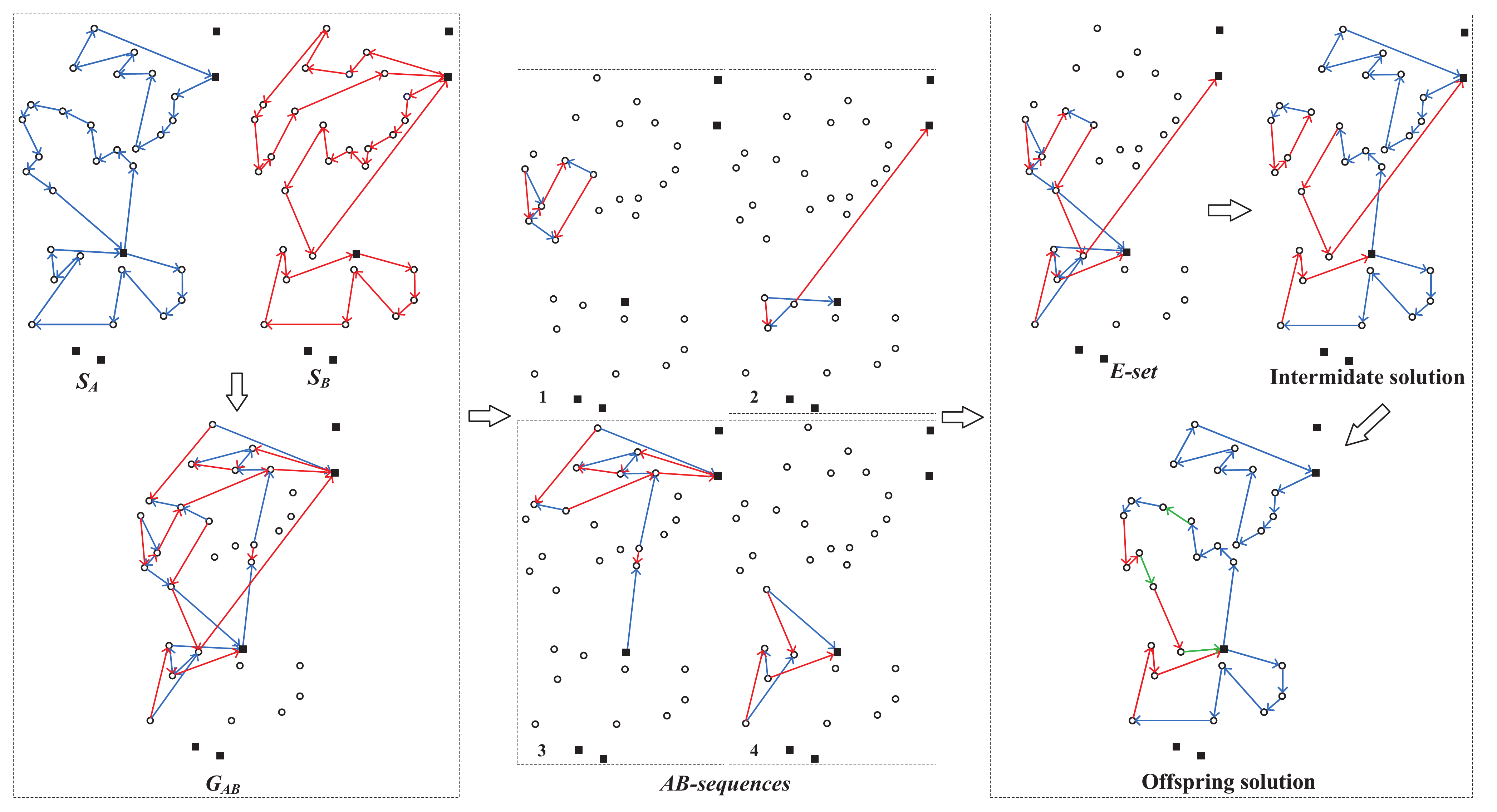}
	\caption{Illustration of the crossover procedure.}
	\label{fig_eax} %% label for entire figure
\end{figure}

Let $S_A$, $S_B$, and $S_C$ be three parent solutions randomly selected in the population. We define their directed graphs, $G_A = \{V,E_A\}$, $G_B=\{V,E_B\}$ and $G_C = \{V,E_C\}$ where $V=D  \cup C$ and $E_X$ ($X=A,B,C$) is the set of directed edges traveled by parent solution $S_X$. Suppose that $S_A$, $S_B$, and $S_C$ are recombined in this order. Then the MPEAX crossover first recombines parents $S_A$ and $S_B$ to get an intermediate offspring solution, which is then recombined with parent $S_C$ to get the final offspring. The specific steps of MPEAX are described as follows and an illustrative example is provided in Fig. \ref{fig_eax}. 

\begin{enumerate}

\item We create a joint graph $G_{AB}=\{V,(E_A \cup E_B) \setminus (E_A \cap E_B)\}$ from $G_A = \{V,E_A\}$ and $G_B=\{V,E_B\}$.

\item The edges in $G_{AB}$ are grouped into AB-\textit{sequences}. An AB-\textit{sequence} begins with a randomly selected node that has connected edges. Then, an adjacent edge is chosen randomly with respect to this node, and edges from $G_A$ and $G_B$ with the same degree for their common node are chosen to be alternately linked. When an edge connecting to the depot is selected, the next chosen edge can be any edge connected to any depot with the same degree. Once an edge is chosen such that it has the same degree as the first selected edge at the first chosen node, an AB-\textit{sequence} is formed. This process is repeated until no edge exists in $G_{AB}$.

\item The \textit{E-set} is built by randomly selecting an AB-\textit{sequence}, and then selecting the AB-\textit{sequences} that share at least one node with the chosen sequence to form the \textit{E-set}. Then take parent $S_A$ as the base solution, remove the edges from $S_A$ and add the edges from $S_B$ included in \textit{E-set}. The step leads to an intermediate solution.
 
\item It is possible that the intermediate solution contains sub-tours (i.e., cycles consisting exclusively of customer nodes). If this happens, they are eliminated by the 2-opt* operator by removing two arcs (one from the sub-tour, one from the existing route) and connecting the sub-tour to the exiting route by adding two new arcs. It is also possible that some routes start and end at different depots, making the route not a cycle. To address this issue, we select the depot that is closer to the first node of the route as the depot for that route. Once this issue is solved, we obtain an intermediate offspring solution from parents $S_A$ and $S_B$, we use $S_O$ to denote this  solution.

\item The last parent solution $S_C$ is then used to be recombined with the intermediate offspring solution $S_O$, following the same procedure used to crossover parents $S_A$ and $S_B$. This leads to the final offspring solution.
\end{enumerate} 

In the example of Fig. \ref{fig_eax}, two out of the five depots (square points) are selected as the opened depots. MPEAX generates four AB-\textit{sequences} (step 2), and AB-\textit{sequence} 4 is selected as the central AB-\textit{sequence}, which shares common nodes with AB-\textit{sequences} 1 and 2. Then the \textit{E-set} consists of three AB-\textit{sequences} (step 3). In the intermediate solution, there is a sub-tour (the small tour), and a tour is not a cycle (involving two depots). After fixing these problems in step 4, three new arcs are introduced, shown in green, leading to the final offspring.

\subsection{Repair procedure}
\label{subsec repair}

%Our approach for handling multiple depots within the MPEAX can potentially result in the selection of more depots than allowed. Additionally, as vehicle capacity constraints are not considered, there may be violations of capacity constraints. To address these issues, we integrated a two-step repair procedure into our proposed algorithm. The first step ensures that the number of opened depots does not exceed the given limit, while the second step focuses on repairing capacity violations.

MPEAX can generate infeasible offspring with more open depots than allowed. In addition, since MPEAX does not take vehicle capacity into account, the capacity constraint may be violated. To address these issues, we use a two-step  procedure to repair an infeasible offspring solution. The first step ensures that the number of open depots doesn't exceed the allowed limit, while the second step focuses on repairing capacity violations.

If the number of open depots exceeds the given limit  $N_d$, we use two repair methods. The first method relies on the frequency information of each depot being selected in the high-quality solutions returned by the local search procedure (RL-SOVND). Among the open depots, the top $N_d$ depots with the highest frequency in the offspring being repaired are retained. The second method randomly selects  $N_d$ depots. Once the open depots are determined, the routes involved in the discarded depots are reassigned using a greedy approach. Each such route is assigned to an open depot such that it is the closest depot to the first node of that route.

To deal with capacity violations, we use the inter-neighborhood operator 2-opt* to reassign customer nodes. we assess a solution using the modified objective function $F$ of Section \ref{subsubsec_sovnd}, with the penalty parameter $\beta$ set sufficiently large to strongly penalize capacity violations. The procedure terminates when a feasible solution is reached or when all neighborhood solutions induced by 2-opt* have been explored. Note that this repair procedure doesn't guarantee the feasibility of capacity. The purpose of the capacity repair is to bring the input solution as close to the feasible space as possible. The RL-SOVND procedure that follows will take care of the infeasibility issue because it examines both feasible and infeasible solutions.  

\subsection{Mutation}
\label{subsec mutation}

After the MPEAX procedure, which is designed to preserve shared edges of the parents that contribute to high quality solutions, there may be a high degree of similarity between the offspring and the parents. To ensure a diversified offspring solution, a mutation is applied, with a probability $m_p$, to modify the offspring with two operators: the depot swap for the depots and the ejection chain for the customers. The depot swap operator selects a depot randomly from the set of unopened depots and uses it to replace a randomly selected open depot. The ejection chain operator randomly selects three customer nodes from different routes and swaps their positions in a cyclic manner. This mutation operation is performed $m_l$ times ($m_l$ is called the mutation length). After the mutation procedure, new edges not present in the parent solutions are introduced, and different depot configurations are explored.

\subsection{Reinforcement learning guided VND with strategic oscillation}
\label{subsec rlsovnd}

Our local search method, which is another critical component in our MA algorithm, is a reinforcement learning guided variable neighborhood descent with strategic oscillation (RL-SOVND). RL-SOVND is characterized by two original features. It explores multiple neighborhoods according to a dynamic order determined by reinforcement learning. To examine candidate solutions, it uses strategic oscillation to consider both feasible and infeasible solutions in a carefully controlled manner. %In the following sections, we first present the two main components of the RL-SOVND procedure (Q-learning driven variable neighborhood descent and strategy oscillation) and then the general RL-SOVND algorithm.

\subsubsection{Rationale and general RL-SOVND framework}

\begin{algorithm}[ht]
    \footnotesize
    \caption{Pseudo-code of RL-SOVND}
    \label{Algo_rlsovnd}
        \Input Input solution $S$, set of neighborhoods, slide window length $u_p$, Q-table $Q$, reward matrix $R$
        \Output Local optimum $S_l$, updated Q-table $Q$, and reward matrix $R$
        
         $I_f, I_i \gets 0$ \mytcp{Set counters on consecutive feasible and infeasible solutions}
         $Improve \gets$ true
         $\beta \gets \frac{f(S)}{\sum \limits_{i \in C }{p_i}}$ \mytcp{Initialize the penalty parameter $\beta$}
         $S_l \gets S$ \mytcp{Record the local optimum solution}
        
        \While{$Improve$}{
             $Improve \gets$ false
             $n_e \gets 0$ \mytcp{Initialize the number of explored neighborhood structures}
             $St.\text{clear}()$ \mytcp{Set the current state of explored neighborhoods}
            
            \While{$n_e < \left|N\right|$}{
                 $N_{\delta} \gets \text{Q-learning}(St, Q, R)$\\
                 $S' \gets S \oplus N_{\delta}$ \mytcp{Perform the first improvement with $N_{\delta}$}
                 $St.\text{append}(N_{\delta})$ \mytcp{Update the state}
                 Update $Q$, $R$ \mytcp{See Section \ref{subsubsec_rl}}
                
                \If{$F(S') < F(S)$}{
                     $S \gets S'$  \mytcp{Accept a better solution under $F$, see Section \ref{subsubsec_sovnd}}
                    
                    \If{\textit{Is\_feasible}($S'$) and $f(S') < f(S_l)$}{
                         $S_l \gets S'$ \mytcp{Update the local optimum solution}
                    }
                    
                     Update $I_i$, $I_f$
                    
                    \If{$I_i > u_p$ or $I_f > u_p$}{
                         Adjust the penalty parameter $\beta$ \mytcp{See Section \ref{subsubsec_sovnd}}
                    }
                    
                     $Improve \gets$ true\\
                     \textbf{break}\\}
                \Else{
                     $n_e \gets n_e + 1$
                     }
                
            }
        }
        
        \Return $S_l$, $Q$, $R$ \mytcp{Return $S_l$, $Q$, and $R$}
\end{algorithm}

RL-SOVND explores multiple neighborhoods sequentially with the VND framework \citep{mladenovic1997variable}, raising the important question of how to determine the order of checking these neighborhoods. Two common strategies for determining this order are the random strategy and the prefixed-order strategy. The random strategy inspects the given neighborhoods in a random order. The prefixed-order approach examines the neighborhoods in a fixed sequence, typically determined according to the computational complexity of the neighborhoods. 

However, the random approach does not differentiate the neighborhood structures and ignores the intrinsic differences between the neighborhoods. On the other hand, when using the predefined approach, one faces the difficulty of determining an appropriate order for neighborhood inspection. In addition, the best order may change during the search process, making it impossible to find an all-time best order. An interesting alternative strategy for neighborhood examination is to determine the order according to the specific problem instance to be solved and the search context of the algorithm. % as in \citep{subramanian2010parallel}.

To do this, we consider the determination of the neighborhood exploration order as a sequential decision-making problem and use reinforcement learning to dynamically make the best possible decision. In particular, RL-SOVND uses the renowned reinforcement learning algorithm Q-learning \citep{watkins1992q}.%RL is based on the idea that an agent learns iteratively through trial-and-error interactions in a dynamic environment \citep{sutton2018reinforcement}. In particular, RL-SOVND uses the renowned reinforcement learning algorithm Q-learning \citep{watkins1992q}.

On the other hand, it is known that visiting infeasible solutions during the search process can be beneficial, as shown in studies on constrained problems \citep{li2024flow, wei2023responsive, zhou2021responsive}. This benefit arises from the increased freedom it provides to visit infeasible solutions, allowing the algorithm to more effectively transition between different feasible search regions via infeasible regions. In RL-SOVND, we use the general strategic oscillation method \citep{glover2011case}, which allows the algorithm to search in both feasible and infeasible regions with a focus on feasible and infeasible boundaries. For this, we devise a mechanism to prevent, based on a penalty parameter $\beta$ and search information, the algorithm from getting stuck in either feasible or infeasible space for too long. 

%To evaluate an infeasible solution, a penalty term is introduced in the extended objective function $F$ to penalize constraint violations (capacity constraint in the LLRP), as shown in Equation \ref{eq penalty objective}. The penalty parameter $\beta$ plays a critical role in balancing the exploration of feasible and infeasible search spaces. In RL-SOVND, we dynamically adjust the penalty parameter based on the search information to prevent the algorithm from getting stuck in either feasible or infeasible space for too long. 

%This approach is consistent with the strategy oscillation \citep{glover2011case}. Another advantage is that this method enables the algorithm obtain more chance to discover possible high-quality solutions located on the boundary between the feasible and infeasible spaces. 

%\begin{center}
%  \begin{equation}
 % \label{eq penalty objective}
%    %f^p(x)=f(x) + \beta \sum_{k=1}^{N_k}{max(0,\sum_{i \in H_k }{p_i}-P)}
%		F(x)=f(x) + \beta \sum_{k=1}^{N_k}{max(0,\sum_{i \in H_k }{p_i}-P)}
%  \end{equation}
%\end{center}

Algorithm \ref{Algo_rlsovnd} shows the general RL-SOVND framework. Initially, the penalty parameter $\beta$ ( Section \ref{subsubsec_sovnd}) is set based on the objective value of the input solution $S$ and the total customer demand $p_i$ (line 3). The algorithm then enters the main loop to improve the current solution. Within this loop, neighborhood structures are systematically explored. The selection of the next neighborhood $N_{\delta}$ to be examined is determined using Q-learning (line 10, Section \ref{subsubsec_rl}). This is done based on the current state $St$ (comprising the explored neighborhood structures) and the historical search information contained in the Q-table $Q$ and the reward matrix $R$. For the chosen neighborhood $N_{\delta}$, the neighborhood solutions are explored using the first improvement strategy (line 11). Subsequently, $St$, $Q$ and $R$ are updated based on the outcome of the executed action (lines 12-13). If an improved solution $S'$ is found in the neighborhood $N_{\delta}$ under the extended objective function $F$ (Section \ref{subsubsec_sovnd}), the current solution $S$ is updated (line 15). Moreover if $S'$ is feasible and is better than the recorded best feasible solution found during the current RL-SOVND run, $S_l$ is also updated by $S'$ (line 17). The counter  for consecutive accepted feasible or infeasible solutions is also updated (line 18), and the penalty parameter $\beta$ is adjusted if the predefined condition is met (lines 19-20). Then, the algorithm returns to the beginning of the main loop (line 22). If no improvement is achieved with the current neighborhood, the next neighborhood structure is explored. The algorithm terminates after exploring all neighborhood structures without improvement, and the best local optimum solution $S_l$, the updated Q-table $Q$ and reward matrix $R$ are returned (line 25).

\subsubsection{Q-learning for deciding the exploring order}
\label{subsubsec_rl}

Q-learning uses the so-called Q-value function to estimate the expected long-term cumulative reward associated with performing a particular action within a given state. We use Q-learning to determine the most suitable neighborhood to explore in order to improve the current solution, considering that some neighborhoods have already been explored. We define the fundamental notations of Q-learning including states, actions, transition policy, and rewards used in our algorithm as follows.
\begin{itemize}

\item States ($ST$): The state set comprises the explored neighborhood structures.

\item Actions ($A$): The set of available actions depends on the current state. In a given state $st$, there is a specific action set denoted as $A(st)$, consisting of the neighborhood structures that have not yet been examined. %An action is to pick an action to performed from the action set.

\item Rewards ($R$): An immediate reward $r=R(st,a)$ is assigned when an action $a \in A(st)$ is executed at the current state $st$. The reward matrix $R$ is updated after the selected neighborhood is examined. Further details about the updating process can be found below.

\item Transition policy: The algorithm employs an $\varepsilon$-greedy policy to govern state transitions between different states. This policy selects with a probability of $\varepsilon$ the action $a^*$ from the action set $A(st)$ of current state $st$ that maximizes the Q-value, i.e., $a^* = \arg\max Q(st, a)$, where $a \in A(st)$. Meanwhile, there is a probability of 1-$\varepsilon$ to randomly choose an action.

%Within this policy, an action $a^*$ that maximizes the Q-value is selected with a probability of $\varepsilon$ from the action set $A(st)$ of current state $st$. It can be expressed as $a^* = \arg\max Q(st, a)$, where $a \in A(st)$. Meanwhile, there is a probability of 1-$\varepsilon$ to randomly choose an action.

\end{itemize}

After the execution of the chosen action, the Q-table is updated according to Equation \ref{eq q_updated}.
 
\begin{center}
  \begin{equation}
  \label{eq q_updated}
    Q(st,a)=(1-\alpha)Q(st,a) + \alpha [R(st,a) + \gamma  \underset{a' \in A(st')} {max} {(Q(st',a'))}]
  \end{equation}
\end{center}

In this equation, $st$ represents the current state, $a$ corresponds to the current action, $st'$ denotes the next state resulting from the current action $a$, and $\alpha$ and $\gamma$, both in the range of [0,1], are the learning rate and discount factor, respectively.
The values stored in $R$ represent the reward values associated with specific actions in a given state. When an action is executed, either an improved solution or a local optimal solution among all the neighbourhood solutions is found, denoted as $S_r$. We use the objective value $f(S_r)$ of $S_r$ to update the reward value associated with the state-action pair. We define two terms: $\Delta_r = f(S_c) - f(S_r)$, which can be either positive or negative, and $\Delta_b = f(S_b) - f(S_r)$, where $S_c$ represents the current solution before executing the selected neighborhood, and $S_b$ is the global best solution found. The update mechanism for the reward values is described by Equation \ref{eq reward}.

\begin{center}
  \begin{equation}
  \label{eq reward}
  R(st,a)= 
  \begin{cases}
         \xi R(st,a) + \Delta_r  + max(0,\Delta_b) e^{\left|N\right|-\left|A(st)\right|} & \text{if } \Delta_r > 0 \\
         \xi R(st,a) + \Delta_r  & \text{if } \Delta_r < 0
    \end{cases}
   \end{equation}
\end{center}

where $\xi$ denotes the discount coefficient, set to 0.95. This coefficient is used to reduce the influence of historical information and to give more weight to recent performance. In particular, an additional bonus reward is given when a new best solution is found, and this reward increases with the number of neighborhoods explored. This is because more reward should be given as high-quality neighborhood solutions become scarcer.

\subsubsection{Variable neighborhood descent with strategic oscillation}
\label{subsubsec_sovnd}

The VND procedure in our algorithm explores seven distinct neighborhood structures induced by the following move operators.

$N_1 \ (Relocate)$. It relocates a customer node from its initial location to another position within the same or a different route.

$N_2 \  (Swap)$. It is associated with both customer nodes and depot nodes. It involves swapping the positions of two nodes, which can be from the same or different routes. The nodes to be swapped must be of the same type, i.e. a customer node can only be swapped with another customer node, and similarly for depot nodes.

$N_3 \ (2\text{-}opt)$. It can be applied to the nodes within the same route (intra-route) or the nodes of different routes (inter-route). The intra-route operator deletes two non-adjacent edges and adds two new edges. Meanwhile, the edges between the deleted edges are reversed. The inter-route operator, also called \textit{2-opt*}, deletes two edges and adds two new edges.

$N_4 \ (2\text{-}relocate)$. It relocates two consecutive customer nodes from their original positions to different locations within the same or different routes.

$N_5 \ (Node\text{-}arc \enspace swap)$. It swaps the positions of a customer node and an arc (two consecutive customer nodes), which can occur within the same route or between different routes.

$N_6 \ (Arc\text{-}arc \enspace swap)$. It swaps two consecutive customer nodes from either the same or different routes.

$N_7 \ (Swap*)$. This is an inter-route operator that selects two customers from different routes, removes them from their original positions, and inserts them into the best position within each other's route. This move operator is only executed  when the routes of the selected customers overlap, following the approach in \cite{vidal2022hybrid}.

It's worth noting that we limit the neighborhood of each customer to include only the $\delta$-nearest vertices, where $\delta < \left|V \right|$. The reason for this is that solutions involving edges with long distances are less likely to be of high quality. This method increases the computational efficiency by avoiding the examination of less promising solutions. Notably, this approach has been shown to be effective in solving other routing problems \citep{helsgaun2000effective, toth2003granular}.

RL-SOVND uses the general strategic oscillation method \citep{glover2011case} to explore both feasible and infeasible solutions within these neighborhood structures. To evaluate an infeasible solution, we define an extended objective function $F$, as shown in Equation \ref{eq penalty objective}, which is a combination of the objective function $f$ and a penalty term to deal with constraint violations. The penalty parameter $\beta$ is used to balance the exploration of feasible and infeasible search spaces and is dynamically adjusted using search information. This helps the algorithm not to get stuck in feasible or infeasible space for too long. 

\begin{center}
  \begin{equation}
  \label{eq penalty objective}
    %f^p(x)=f(x) + \beta \sum_{k=1}^{N_k}{max(0,\sum_{i \in H_k }{p_i}-P)}
		F(x)=f(x) + \beta \sum_{k=1}^{N_k}{max(0,\sum_{i \in H_k }{p_i}-P)}
  \end{equation}
\end{center}

Specifically, the VND procedure maintains a sliding window of length of $u_p$ iterations to evaluate the feasibility of the accepted solutions within the window. If all accepted solutions are feasible, we decrease the penalty parameter to promote exploration of infeasible spaces. Conversely, if all solutions in the window are infeasible, we increase the penalty parameter to encourage the algorithm to explore feasible spaces. If both feasible and infeasible solutions are accepted within the window, we keep the penalty parameter unchanged. The specific method for adjusting this parameter is shown in Equation \ref{eq beta} where $I_i$ is the number of accepted infeasible solutions in the sliding window, $I_f$ is the number of accepted feasible solutions, $u_p$ is the predefined threshold for adjusting $\beta$, and $rand(0,1)$ is a random number 0 or 1.

%The solutions encountered during the local search procedure are evaluated using Equation \ref{eq penalty objective}. As our algorithm allows exploration of the infeasible search space, it is crucial to prevent the algorithm from remaining in either the feasible or infeasible search space for an extended duration. The penalty parameter $\beta$, which penalizes constraint violations, plays a vital role in balancing the exploration between the feasible and infeasible search spaces. In our algorithm, we dynamically adjust the penalty parameter based on short-term search information. 

\begin{center}
  \begin{equation}
  \label{eq beta}
  \beta =
    \begin{cases}
        \beta (1.5 + rand(0,1)) & \text{if } I_i = u_p \\
        \frac{\beta}{1.5 + rand(0,1)} & \text{if } I_f = u_p
    \end{cases}
  \end{equation}
\end{center}

\subsection{Population updating}
\label{subsec population}

Population updating aims at maintaining an appropriate diversity among the solutions in the population. The updating mechanism used takes into account both the quality of the solution and its contribution to the population diversity. The contribution to diversity is assessed by measuring the distance between the new solution and the population. %We employ this method to determine whether a new solution from the local search procedure (RL-SOVND) should be added to the population. 

Given two solutions, $S_a$ and $S_b$, their distance is the number of non-common edges between the solutions, which is determined by Equation \ref{eq distance}, where $E$ represents the edge set of a solution. Accordingly, the distance between a solution and the population is defined as the minimum distance between this solution and any solution from the population (excluding itself if it is also part of the population), as shown in Equation \ref{eq Pdistance}.

\begin{center}
  \begin{equation}
  \label{eq distance}
  IDist(S_a, S_b) = \left|E_a\right|-\left|E_a \cap E_b\right|   
  \end{equation}
\end{center}

\begin{center}
  \begin{equation}
  \label{eq Pdistance}
  PDist(S, Pop) = min\{IDist(S,S_i):S_i \in Pop \setminus S\}  
  \end{equation}
\end{center}

We employ this method to determine whether a new solution from the local search procedure (RL-SOVND) should be added to the population. We first check if there is a clone of the new solution in the population (i.e., $PDist(S,Pop) = 0$). If this is the case, we discard the new solution. Otherwise, we add the new solution into the population, resulting in a modified population called $Pop'$. Next, we re-evaluate the fitness of all solutions in $Pop'$ using their quality and distance to this population by Equation \ref{eq fitness}, and the solution with the worst fitness value is removed from the population. In this equation, a normalization is applied since the quality and distance values are not of the same dimension. We define $f_{max} = \max\{f(S_i):S_i \in Pop'\}$ and $f_{min} = \min\{f(S_i):S_i \in Pop'\}$ as the maximum and minimum objective values within the population $Pop'$. Additionally, $PD_{max} = \max\{PDist(S_i, Pop'):S_i \in Pop'\}$ and $PD_{min} = \min\{PDist(S_i, Pop'):S_i \in Pop'\}$ represent the maximum and minimum distances between the solutions and the population. The parameter $\psi$ is empirically set to 0.55.

\begin{center}
  \begin{equation}
  \label{eq fitness}
  fit(S)= \psi \frac{f_{max}-f(S)}{f_{max}-f_{min}} + (1-\psi) \frac{PDist(S,P')-PD_{min}}{PD_{max}-PD_{min}}
  \end{equation}
\end{center}

If the best solution found so far is not updated during $I_r$ (set to 1000) consecutive generations, indicating a search stagnation, we introduce diversity into the population to facilitate escape from deep local optima. First, we randomly remove half of the solutions in the population, while keeping the best solution. Then, for the given population size, new solutions are added either using the random initialization method (Section \ref{subsec initialization}) or by randomly selecting solutions from an adaptive memory $M$. The adaptive memory stores the most recent 3000 local optima found by the local search procedure RL-SOVND. To introduce more diversity into the population, only solutions in the first half of $M$ are considered, corresponding to the solutions added to the memory earlier.  %In our algorithm, we set the parameter $I_r$ to be 1000.

\subsection{Discussions}
Our RLHEA algorithms has a number of novelties compared to the existing methods. 

The MPEAX crossover is derived from the dEAX crossover of the two-individual evolutionary algorithm (TIEA) for the CCVRP and MDCCVRP \citep{zou2024two}, which can be regarded as special cases of the LLRP. MPEAX is divided into two phases, each with two parent solutions. For each phase, MPEAX and dEAX share the same operations for the first three steps, while the remaining two steps are different. In fact, since the number of open depots in the LLRP is limited, the intermediate solution (step 4) may require solution repair by selecting the given number of open depots. 

In addition, MPEAX extends the dEAX crossover by incorporating three parent solutions to mitigate the loss of diversity resulting from considering the direction of edges during the crossover procedure. When using three parent solutions, the crossover order of the three parents needs to be carefully considered. Indeed, compared to the first two parents, the features of the last parent are only diluted once by crossover, increasing the opportunity of transmitting its edges to the offspring solution. For MPEAX, based on this observation, we select the individual with the shortest life in the population (inserted into the population last) as the third parent, which effectively enhances the diversity of the population. 

Compared to the work \citep{moshref2016latency}, which uses a simple OX crossover applied to a giant tour, the MPEAX crossover allows the offspring to naturally inherit the common edges from the parents. This is beneficial for creating more promising offspring and thus increasing the search efficiency of the algorithm. 

Finally, our local search procedure, RL-SOVND, follows the general VND framework, which relies on multiple neighborhoods. This raises the critical issue of determining the best exploration order of the adopted neighborhoods. Compared to other methods that also use VND to solve routing problems \citep{ren2023effective,zou2024two}, our RLHEA algorithm differs in its approach to learning the best neighborhood exploration order from the search information. Indeed, most existing methods explore their neighborhoods in a fixed or random order. In contrast, RLHEA uses Q-learning to dynamically determine the best order for neighborhood exploration. This adaptive approach makes RLHEA more effective and enhances its performance, as demonstrated in Section \ref{subsec benefits q-learning}. And it can be applied to any local search algorithm involving a portfolio of neighborhoods or search operators.

\section{Computational results}
\label{sec computation}

We now present an extensive computational evaluation of the RLHEA algorithm on the benchmark instances for the LLRP and a comparison with the state-of-the-art algorithms.

\subsection{Benchmark instances}
\label{subsubsec benchmark}

We use three sets of 76 benchmark instances introduced by \cite{moshref2016latency}. %These instances are widely used for different variants of the LRP. The original data for the three sets are available online\footnote{http://prodhonc.free.fr/}. In the LLRP problem, vehicles are used greedily, which requires the specification of the number of available vehicles and open depots. These values are set based on those obtained in \cite{vincent2010simulated}.

\textit{Set Tuzun-Burke}: This dataset consists of 36 instances with 100 to 200 customers and 10 to 20 depots, and is considered as the most challenging of the benchmark instances. No optimal solutions have been reported in this set. 

\textit{Set Prodhon}: This dataset contains 30 instances with 20 to 200 customers and 5 to 10 depots. 11 instances have been solved optimally in the literature. 

\textit{Set Barreto}: This dataset consists of 10 instances with 21 to 134 customers and 5 to 14 depots. Six instances with less than 50 customers have been solved optimally in the literature.

\subsection{Experimental conditions and reference algorithms}
\label{subsec Experimental protocol}

The RLHEA algorithm has the following main parameters: mutation probability $m_p$, mutation length $m_l$, learning rate $\alpha$, discount factor $\gamma$, greedy probability $\varepsilon$, length of the sliding window $u_p$, and neighborhood reduction parameter $\delta$. To determine suitable values for these parameters, we employed the automatic parameter tuning package Irace \citep{lopez2016irace}. Through the tuning process, we obtained the configuration presented in Table \ref{table para tuning}. This configuration represents the default parameter setting for our algorithm. Moreover, RLHEA uses a population of 20 individuals.

\begin{table}[ht]
\renewcommand\tabcolsep{2.0pt}
\renewcommand{\baselinestretch}{1}\small\normalsize
\tiny
\begin{center}
\centering
\caption{Parameter tuning results}
\label{table para tuning}
\begin{tabular}{lrrrr}
\toprule
Parameter         & Related section                     & Description                               &Considered values        &Final values  \\
\midrule
$m_p$            &\ref{subsec mutation}                      &mutation probability                      &\{0,0.1,0.2,0.3\}          &0.1   \\
$m_l$            &\ref{subsec mutation}                     &mutation length                           &\{1,2,3,4,5\}              &2  \\
$\alpha$         &\ref{subsubsec_rl}                &learning rate                     &\{0.1,0.2,0.3,0.4,0.5\}         &0.2  \\
$\gamma$         &\ref{subsubsec_rl}                &discount factor                     &\{0.8,0.85,0.9,0.95\}         &0.85  \\
$\varepsilon$    &\ref{subsubsec_rl}                &probability of $\varepsilon$-greedy  &\{0.7,0.75,0.8,0.85,0.9,0.95\}  &0.7   \\
$u_p$            &\ref{subsubsec_sovnd}                  &length of the slide window   &\{2,4,6,8,10\}                       &4   \\
$\delta$         &\ref{subsubsec_sovnd}             &granularity threshold        &\{10,15,20,25,30\}                     &20  \\
\bottomrule
\end{tabular}
\end{center}
\end{table}

According to the literature review in Section \ref{sec Introduction}, four studies have addressed the LLRP problem. The earliest algorithm MA \citep{moshref2016latency} retains only few best-known solutions. Consequently, we have excluded it from our comparative study. The reference algorithms for comparison include GBILS proposed in \citep{nucamendi2022new}, three algorithms (SA-VND0, SA-VND1, SA-AND2) introduced in \citep{osorio2023effective}, and the M-ILS algorithm presented in \citep{osorio2023iterated}. The M-ILS algorithm has two versions, one yielding superior results with 30 runs and another with 5 runs; in our study, we exclusively compared with the former (with 30 runs). For the \textit{Set Tuzun-Burke}, there is no result reported for the GBILS algorithm. Among these algorithms, M-ILS stands out as the most powerful, retaining almost all of the current best-known solutions.

Our RLHEA algorithm was programmed in C++  and compiled using the g++ 10.2.1 compiler with the -O3 optimization option. The experiments were conducted on a Xeon E5-2670 processor operating at 2.5GHz with 2GB RAM, running Linux with a single thread. Our algorithm was executed 30 times for each instance, following the approach employed by M-ILS. The stopping condition for our algorithm was set to 5000 generations (crossovers). For the three algorithms SA-VND0, SA-VND1, SA-VND2, and the M-ILS algorithm, the authors generously provided the C++ source codes that we ran on our computer, making it possible to perform a fair comparative study. The algorithm parameters and stopping conditions are set to match the criteria outlined in the original paper.  The tested instances and the best solutions achieved by our algorithm are accessible online\footnote{https://github.com/YujiZou/LLRP}, and the code of our RLHEA algorithm will also be made publicly available. 

\subsection{Computational results and comparison}
\label{subsec computation result}

%We present a comparison of the results obtained by our RLHEA algorithm with those from the reference algorithms. 

Table \ref{tab summarized result} provides a summary of the comparative results on the three datasets between our RLHEA algorithm and the reference algorithms, while Tables \ref{table results tuzun}--\ref{table results barreto} of the Appendix show the detailed results, including the best and average objective values as well as the average CPU running time. In Table \ref{tab summarized result}, the first column represents the dataset. Columns  $f_{best}$ and $f_{avg}$ provide a summary in terms of the best and average objective values achieved among 30 independent runs. The column labeled "$\#$Wins" indicates the number of instances where RLHEA outperformed the reference algorithm, "$\#$Ties" shows the number of instances with equal results, and "$\#$Losses" indicates the number of instances where RLHEA performed worse than the reference algorithm. The \textit{p}-values from the Wilcoxon signed-rank test (with a significance level of 0.05) applied to the best and average values are also indicated, verifying the statistical significance of the performance differences between RLHEA and each reference algorithm. "BKS" represents the best-known solutions ever reported so far in the literature.

The results in Table \ref{tab summarized result} show that our RLHEA algorithm is highly competitive compared to the reference algorithms in the best and average objective values. Overall, RLHEA achieved new best-known solutions for 51 (out of 76) instances and matched all best-known results for the remaining instances (with no worse results). Specifically, for the most challenging set \textit{Tuzun-Burke}, our algorithm discovered 31 new record-breaking solutions out of the 36 instances. For the set \textit{Prodhon}, RLHEA reported 18 new best results out of the 19 instances whose optimal solutions were unknown, and for the set \textit{Barreto}, it reached new best results for 2 out of the 4 instances with unknown optimal solutions. Regarding the average objective value, RLHEA outperforms the reference algorithms in all instances in the set \textit{Tuzun-Burke}. For the set \textit{Prodhon} and the set \textit{Barreto}, RLHEA also reports many better average results with no worse results. The \textit{p}-values for $f_{best}$ and $f_{avg}$, excluding the set \textit{Barreto} due to the small sizes of the instances, are all less than 0.05. %These results confirm the superiority of our algorithm over the reference algorithms in terms of both the best and average objective values. %These observation can significantly prove the supriority of our proposed algorithm.

\begin{table*}[ht]
\renewcommand\tabcolsep{3pt}
\tiny
\begin{center}
\centering
\caption{Summarized comparison results of RLHEA against the reference algorithms in terms of the best and average objective values on the three sets of 76 LLRP instances.}
\label{tab summarized result}

\begin{tabular}{llrrrrrrrr}
\toprule
\multirow{2}{*}{Instance}  &\multirow{2}{*}{ Pair algorithms}  &\multicolumn{4}{c}{$f_{best}$}                    &\multicolumn{4}{c}{$f_{avg}$}  \\
                                                               \cmidrule(r){3-6}                                   \cmidrule(r){7-10}
                            &                     & \#Wins   &\#Ties   &\#Losses   & \textit{p}-value             & \#Wins   &\#Ties   &\#Losses   & \textit{p}-value \\
\midrule
\multirow{5}{*}{Tuzun-Burke}  &RLHEA vs. BKS       &31         & 5        & 0        &1.32e-5                          &-         & -       & -       & - \\
                              &RLHEA vs. MILS      &32         & 4        & 0       &9.0e-6                            &36         & 0        & 0        &2.84e-6  \\
                              &RLHEA vs. SA-VND0   &36         & 0        & 0        &2.84e-6                          &36         & 0        & 0        &1.53e-6\\
                              &RLHEA vs. SA-VND1   &35        & 1         & 0        & 3.56e-6                         &36         & 0        & 0        &1.41e-6 \\
                              &RLHEA vs. SA-VND2   &35         & 1        & 0       &3.86e-6                           &36         & 0         &0        &1.53e-6  \\

\hline
\multirow{6}{*}{Prodhon}      &RLHEA vs. BKS       &18         & 12        & 0        &1.96e-4                            &-         & -       & -       &-  \\
                              &RLHEA vs. MILS      &19         & 11        & 0       &1.32e-4                           &27         &3        & 0        &5.61e-6  \\
                              &RLHEA vs. SA-VND0   &21         & 9       & 0        &5.96e-5                               &26         & 4        & 0        &8.29e-6\\
                              &RLHEA vs. SA-VND1   &21         & 9        & 0        &5.96e-5                              &26         & 4        & 0        &8.30e-6 \\
                              &RLHEA vs. SA-VND2   &21         & 9        & 0       &5.96e-5                            &26         & 4         & 0        &8.30e-6  \\
                              &RLHEA vs. GBILS     &26         & 4        & 0       &8.30e-6                            &-         & -        & -        &- \\
                            
\hline
\multirow{6}{*}{Barreto}      &RLHEA vs. BKS       &2         &8        & 0        &0.18                             &-         & -       & -       &-  \\
                              &RLHEA vs. MILS      &3         & 7        & 0       &0.11                          &8         & 2        & 0        &0.01  \\
                              &RLHEA vs. SA-VND0   &3         &7        & 0        &0.11                            &7         & 3        & 0       &0.02\\
                              &RLHEA vs. SA-VND1   &4        &6        & 0        &0.07                              &7         & 3        &0        &0.02 \\
                              &RLHEA vs. SA-VND2   &4         &6        & 0       &0.07                           &6         & 4         &0        &0.03  \\
                              &RLHEA vs. GBILS      &3         & 5        & 0       &0.11                        &-         & -       & -        &-  \\
\bottomrule
\end{tabular}
\end{center}
\end{table*}

\subsection{Assessment of computational efficiency}
\label{subsec ttt}

From the detailed results of Tables \ref{table results tuzun}--\ref{table results barreto}, we observe that our algorithm exhibits significant competitiveness in running time compared to the leading algorithms SA-VND0, SA-VND1, SA-VND2, and M-ILS. To further demonstrate the effectiveness of our algorithm, we conducted a Time-to-Target analysis (TTT) \citep{aiex2007ttt}. This analysis measures the time required for each algorithm to achieve a solution with an objective value at least as good as a predefined target objective value. The TTT presents the empirical probability distributions within the given time to reach the target value. In our TTT analysis, we performed each algorithm (with the source code) 100 times on different instances, recording the time taken to reach the target value. Subsequently, we sorted the times in ascending order and calculated the probability $\rho_i = (i - 0.5)/100$ for each time $T_i$, where $T_i$ represents the \textit{i}th smallest time. %For a more in-depth understanding of the TTT methodology, please refer to \cite{aiex2007ttt}.

\begin{figure}[htbp]
\centering

\begin{minipage}[b]{0.49\textwidth}
  \centering
  \subfloat[122122 (Target value=3850)]{\label{fig ttt122122} \includegraphics[width=\textwidth]{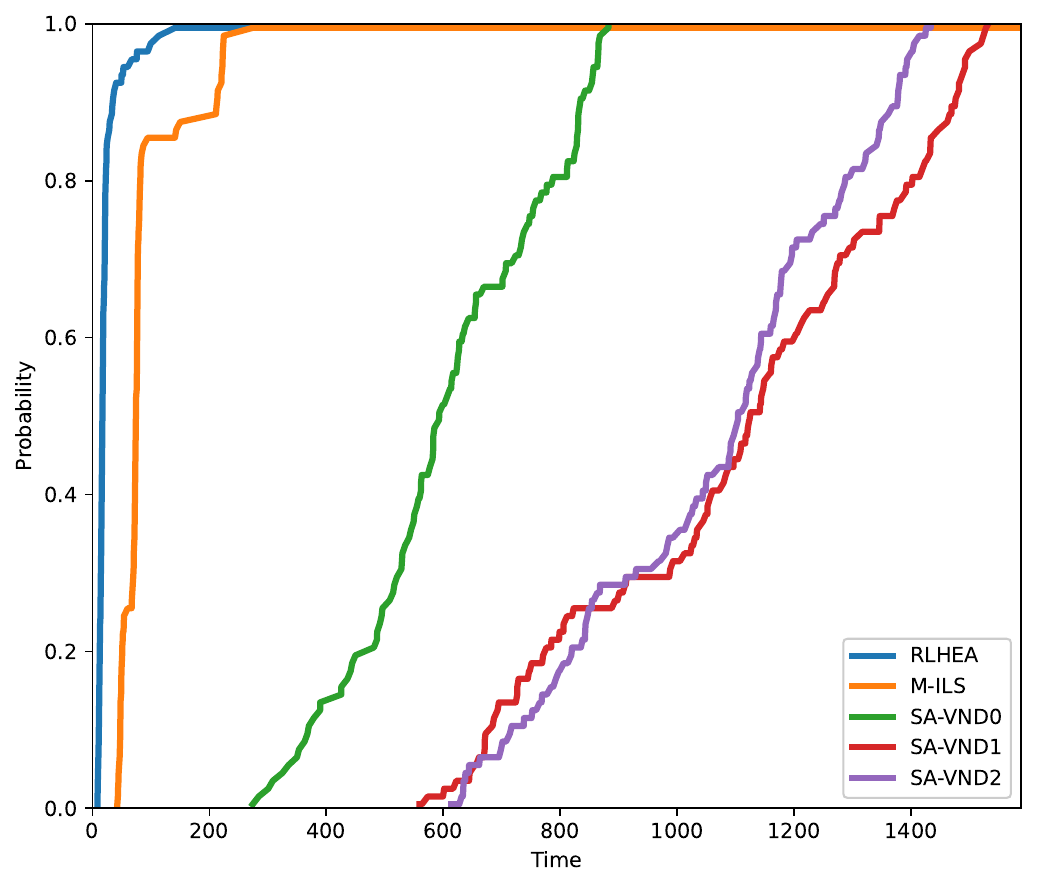}}
\end{minipage}
\begin{minipage}[b]{0.49\textwidth}
  \centering
  \subfloat[123112 (Target value=5150)]{\label{fig ttt123112} \includegraphics[width=\textwidth]{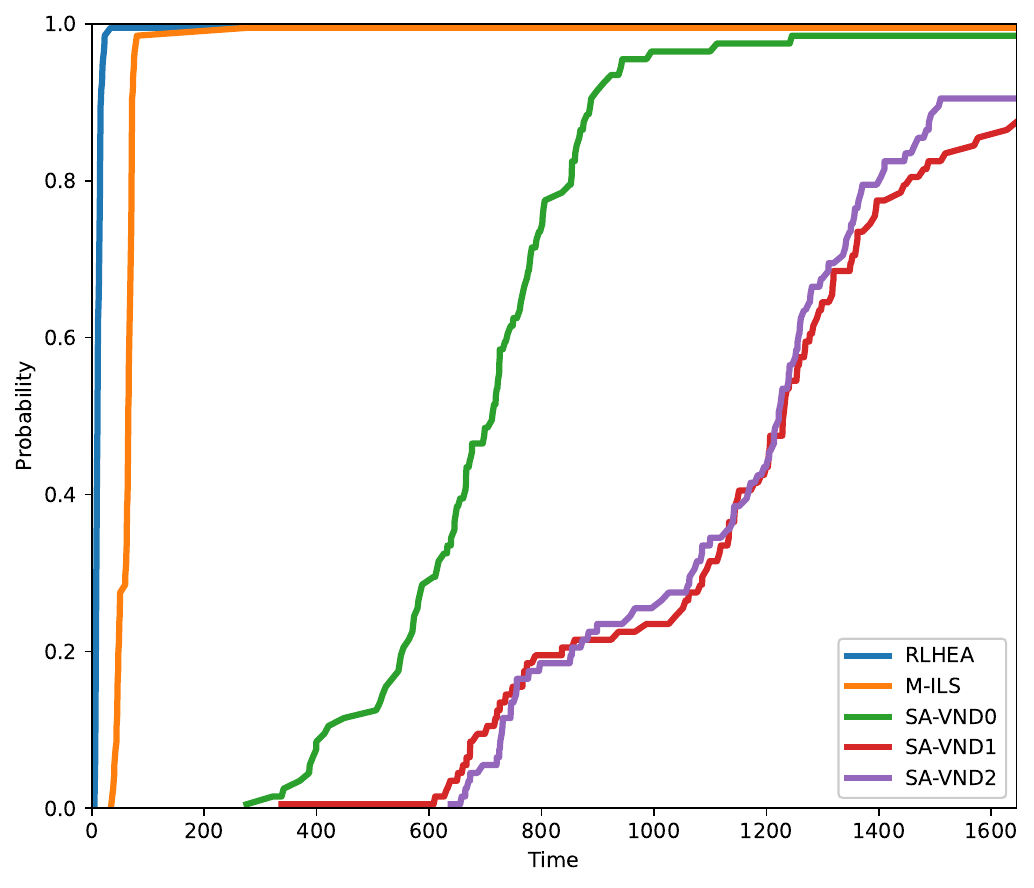}}
\end{minipage}

\begin{minipage}[b]{0.49\textwidth}
  \centering
  \subfloat[123212 (Target value=5300)]{\label{fig ttt123212} \includegraphics[width=\textwidth]{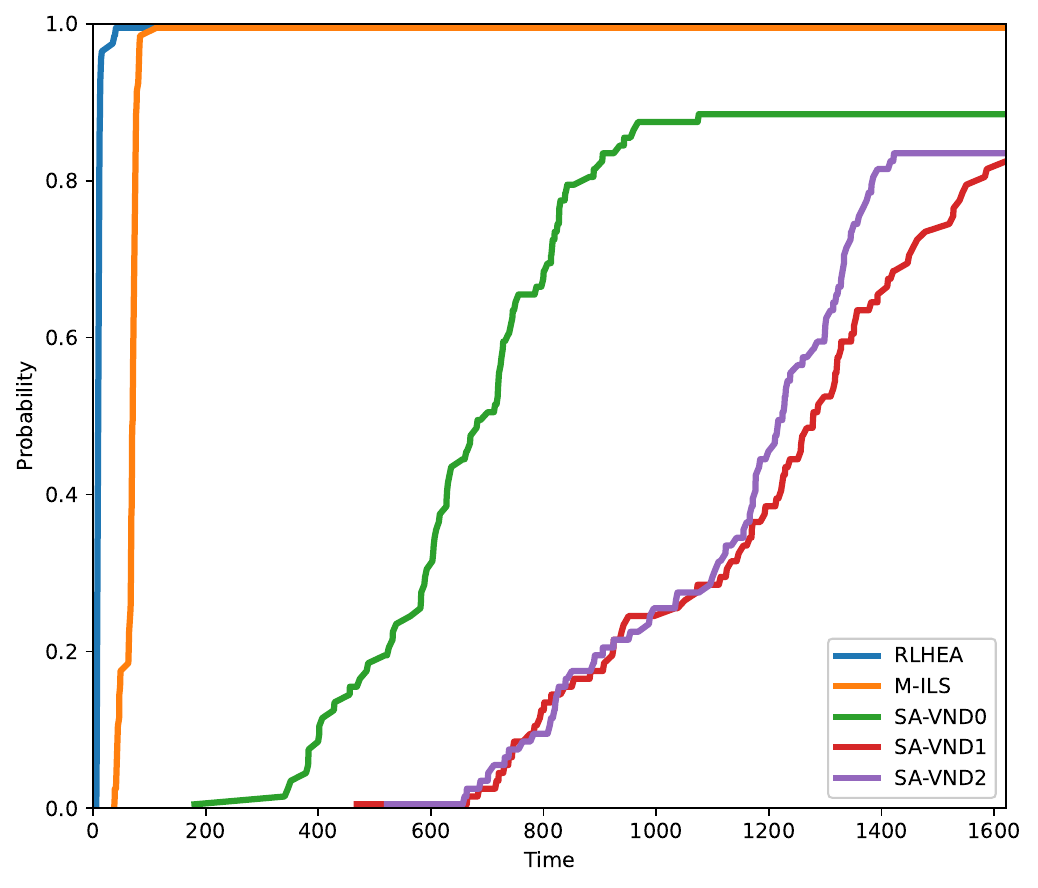}}
\end{minipage}
\begin{minipage}[b]{0.49\textwidth}
  \centering
  \subfloat[200-10-1b (Target value=3500)]{\label{fig ttt200-10-1b}  \includegraphics[width=\textwidth]{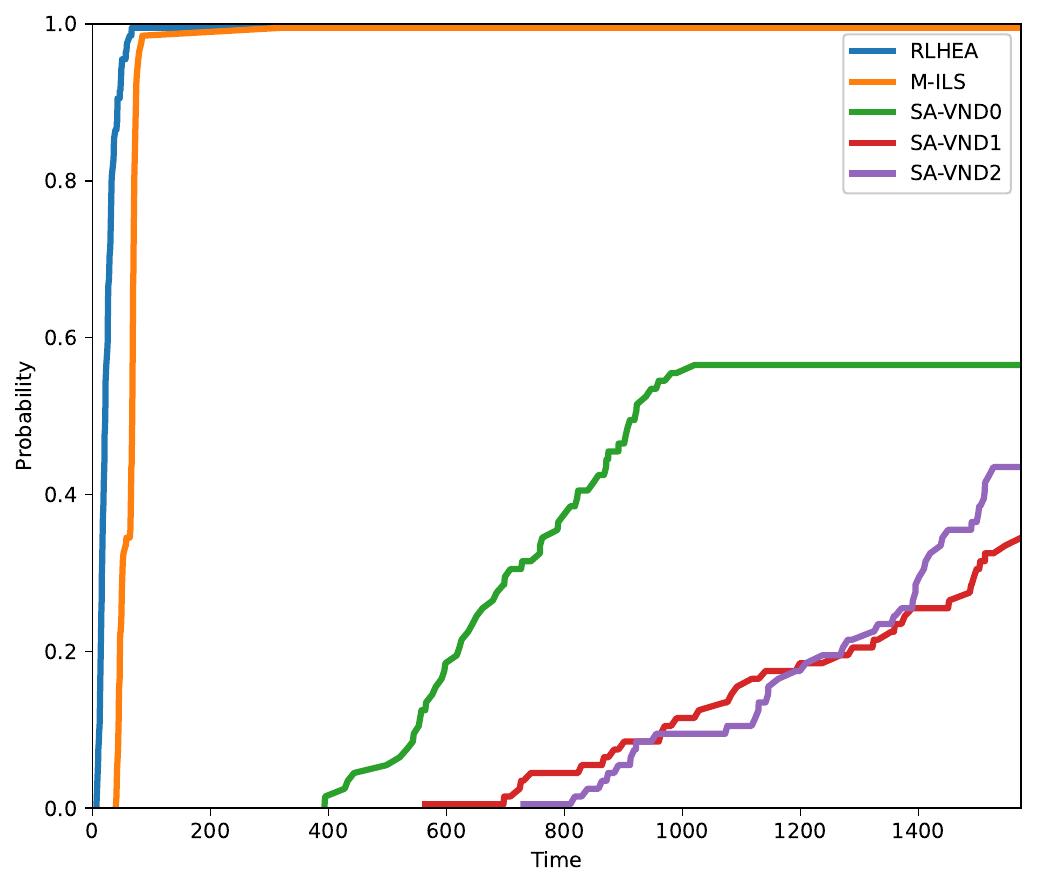}}
\end{minipage}

\caption{Cumulative probability distribution for the time to reach a target value.}
\label{fig ttt}
\end{figure}

Fig. \ref{fig ttt} illustrates the TTT plots for four large instances (122122, 123112, 123212, 200-10-1b) from the set \textit{Tuzun-Burke} and set \textit{Prodhon}. The x-axis represents the time needed to reach the target value, while the y-axis represents the cumulative probability $\rho_i$ of reaching the given target value. The figures show that the TTT curves of our algorithm are consistently above the curves of the reference algorithms, indicating that our algorithm always has a higher probability of reaching the given target value within the same running time. This experiment shows the competitiveness  of RLHEA with state-of-the-art algorithms in terms of computational and search efficiency.

\section{Analysis}
\label{sec analyze}

In this section, we conduct additional experiments to gain deeper insights into the individual influences of the main components of the RLHEA algorithm. We focus on the critical components: the MPEAX crossover, the Q-learning method and the strategic oscillation method. 

\subsection{Rationale behind the MPEAX crossover}

Previous studies on the TSP \citep{nagata2013powerful}, the VRP \citep{arnold2019makes}, and their variants  \citep{he2023general} have revealed that high-quality solutions in these problems often share many common edges, which are likely to be part of the optimal solution. We show experimentally that this is also true for the LLRP, which provides a basis for the MPEAX crossover. Indeed, like the EAX crossover for the TSP, the MPEAX crossover takes advantage of this property by transferring common edges from parent solutions to the offspring, while introducing new edges to increase the diversity of the offspring.

\begin{figure}[htbp]
\centering
\begin{minipage}[b]{0.49\textwidth}
  \centering
  \subfloat[Solution pair shared edge heatmap]{\label{fig influence mpeax subfig1}  \includegraphics[width=\textwidth]{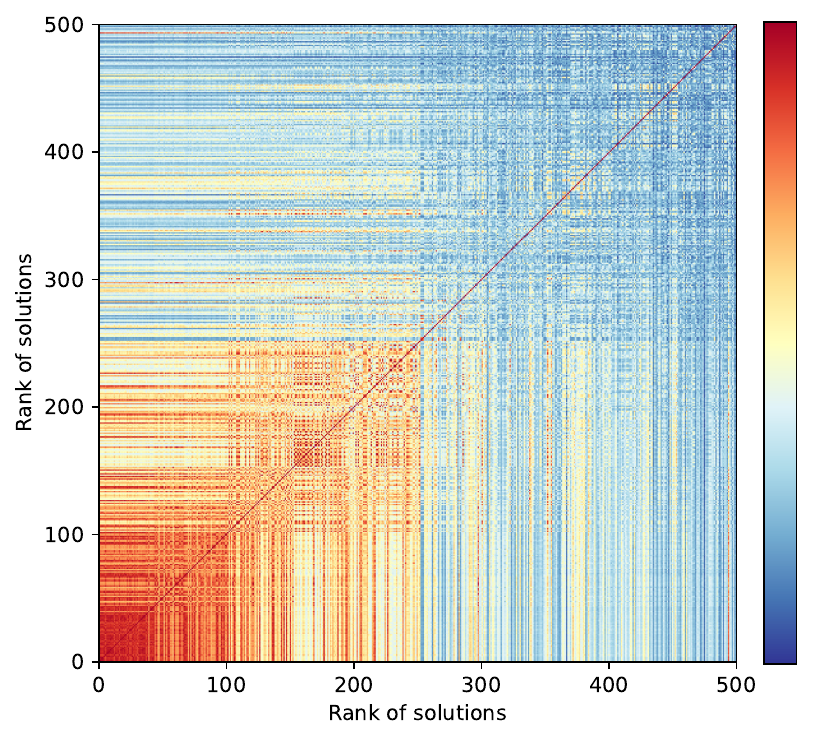}}
\end{minipage}
\hspace{0.000cm}
\begin{minipage}[b]{0.46\textwidth}
  \centering
  \subfloat[Edge sharing ratio plot]{\label{fig influence mpeax subfig2}  \includegraphics[width=\textwidth]{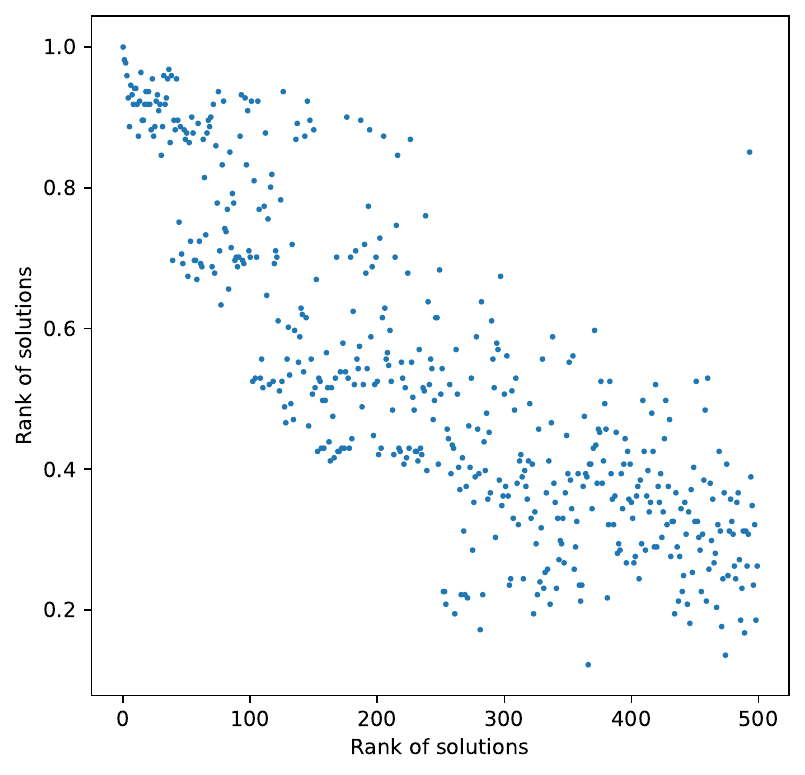}}
\end{minipage}
\caption{The heatmap for the number of shared edges of solution pairs and the scatter plot for edge sharing ratio between solutions and the best-known solution on instance 121222.}
\label{fig heatmap 121222}
\end{figure}

\begin{figure}[htbp]
\centering
\begin{minipage}[b]{0.49\textwidth}
  \centering
  \subfloat[Solution pair shared edge heatmap]{\label{fig influence mpeax subfig1}  \includegraphics[width=\textwidth]{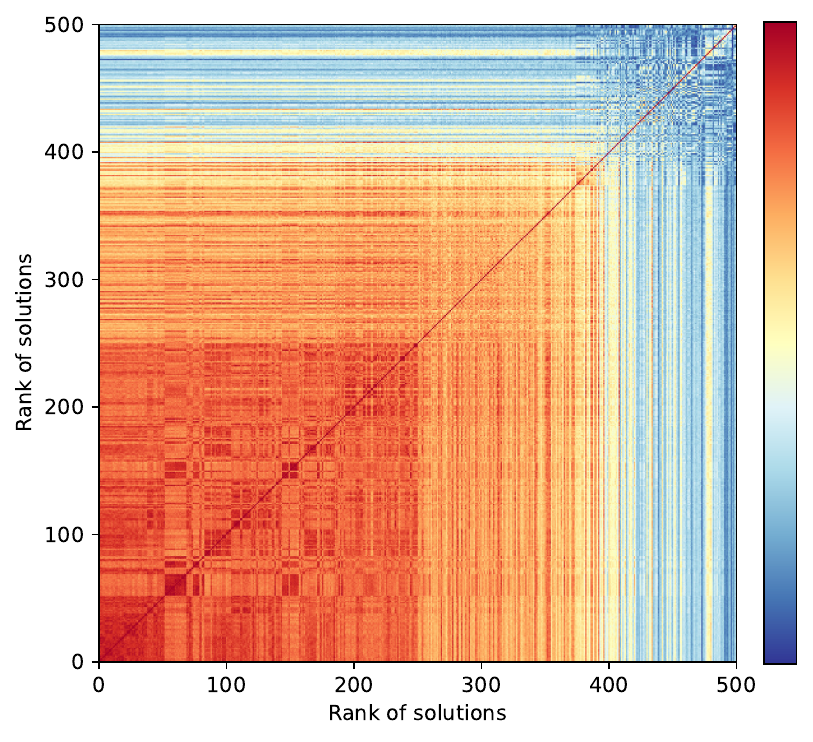}}
\end{minipage}
\hspace{0.000cm}
\begin{minipage}[b]{0.46\textwidth}
  \centering
  \subfloat[Edge sharing ratio plot]{\label{fig influence mpeax subfig2}  \includegraphics[width=\textwidth]{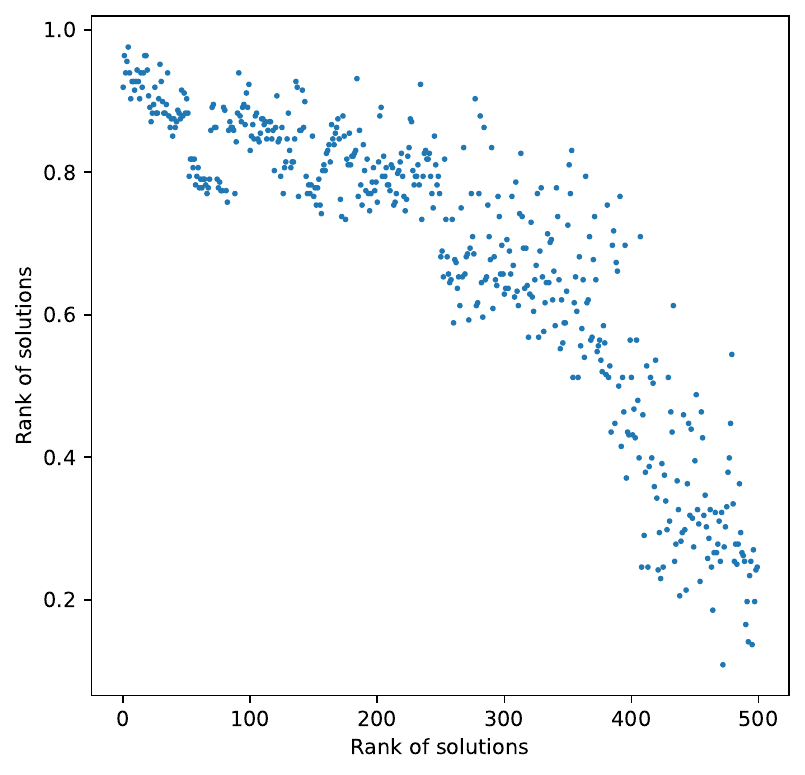}}
\end{minipage}
\caption{The heatmap for the number of shared edges of solution pairs and the scatter plot for edge sharing ratio between solutions and the best-known solution on instance 200-3-1.}
\label{fig heatmap 200-3-1}
\end{figure}

For this study, we focus on two representative instances 121222 and 200-3-1. We ran RLHEA to solve each instance 50 times and collected 15 distinct high-quality local optimal solutions in each run, resulting in a total of 750 unique local optimal solutions per instance. We sorted these 750 solutions in increasing order of  their objective values and selected the top 250 (best) solutions and the 250 worst solutions to form the final set of 500 solutions. We then calculated the number of common edges for each pair of solutions and presented the results as a heat map, as shown in Fig. \ref{fig heatmap 121222}(a) and Fig. \ref{fig heatmap 200-3-1}(a). The x-axis and y-axis represent the rank of the solutions in the solution set, and the color represents the number of common edges. A color closer to red indicates more common edges, while a color closer to blue indicates fewer common edges. To further illustrate this property, in Fig. \ref{fig heatmap 121222}(b) and Fig. \ref{fig heatmap 200-3-1}(b) we show the percentage of edges that a solution $S$ shares with the best solution, where the percentage is calculated by $\frac{\left|E_b \cap E_s \right|}{\left| E_s \right|}$, where $E_b$ is the edge set of the best solution and $E_s$ is the edge set of the solution $S$.

The heatmaps of the two studied instances exhibit the same trend, we can clearly see that the lower-left corner, where the shared edges between high-quality solution pairs are shown, is colored with deep red. The top-right is colored with blue, indicating that solution pairs with poor objective values share fewer edges. Additionally, in the figure showing the relationship between the objective value and the number of shared edges with the best solution, we observe the trend that solutions with higher objective values share more edges with the best solution.

We can conclude that in the LLRP, high-quality solutions also share a high number of common edges, which provides a foundation for the MPEAX crossover to inherit common edges from the parents during the crossover process.

\subsection{Benefits of the MPEAX crossover}
\label{subsec benefits mpeax}

As presented in Section \ref{subsec mpeax}, RLHEA's MPEAX crossover uses three parent solutions to address the problem of diversity degradation due to edge orientation considerations in the crossover process. To study the benefits of this method, we created two algorithm variants, RLHEA1 and RLHEA2. In RLHEA1, we replaced MPEAX with the order crossover used in \cite{moshref2016latency} for the LLRP. In RLHEA2, we applied the MPEAX crossover to only two parent solutions. The remaining components for the two algorithm variants are identical to RLHEA, including the stopping condition set to 5000 generations. We ran the algorithm variants on the large instances with at least 100 customers from the sets \textit{Tuzun-Burke} (36 instances) and \textit{Prodhon} (18 instances). %Detailed information about the instances can be found in Section \ref{sec detailed results}.

\begin{figure}[htbp]
\centering
\begin{minipage}[b]{0.49\textwidth}
  \centering
  \subfloat[Best objective value]{\label{fig influence mpeax subfig1}  \includegraphics[width=\textwidth]{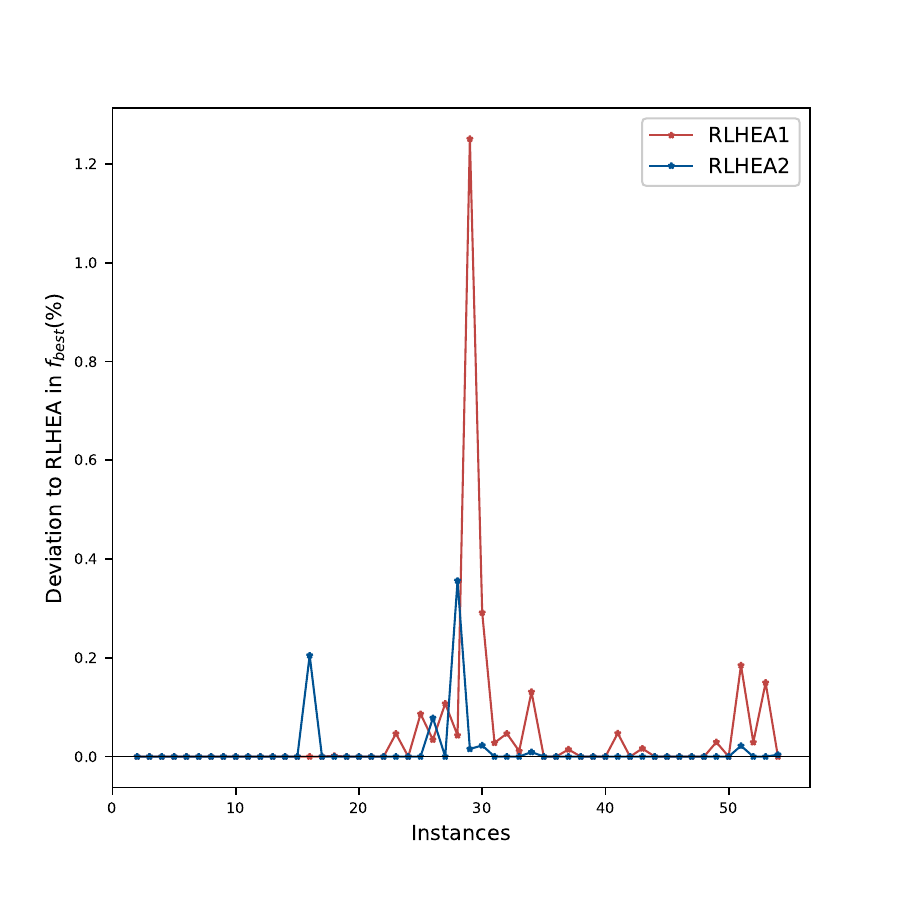}}
\end{minipage}
\hspace{0.000cm}
\begin{minipage}[b]{0.49\textwidth}
  \centering
  \subfloat[Average objective value]{\label{fig influence mpeax subfig2}  \includegraphics[width=\textwidth]{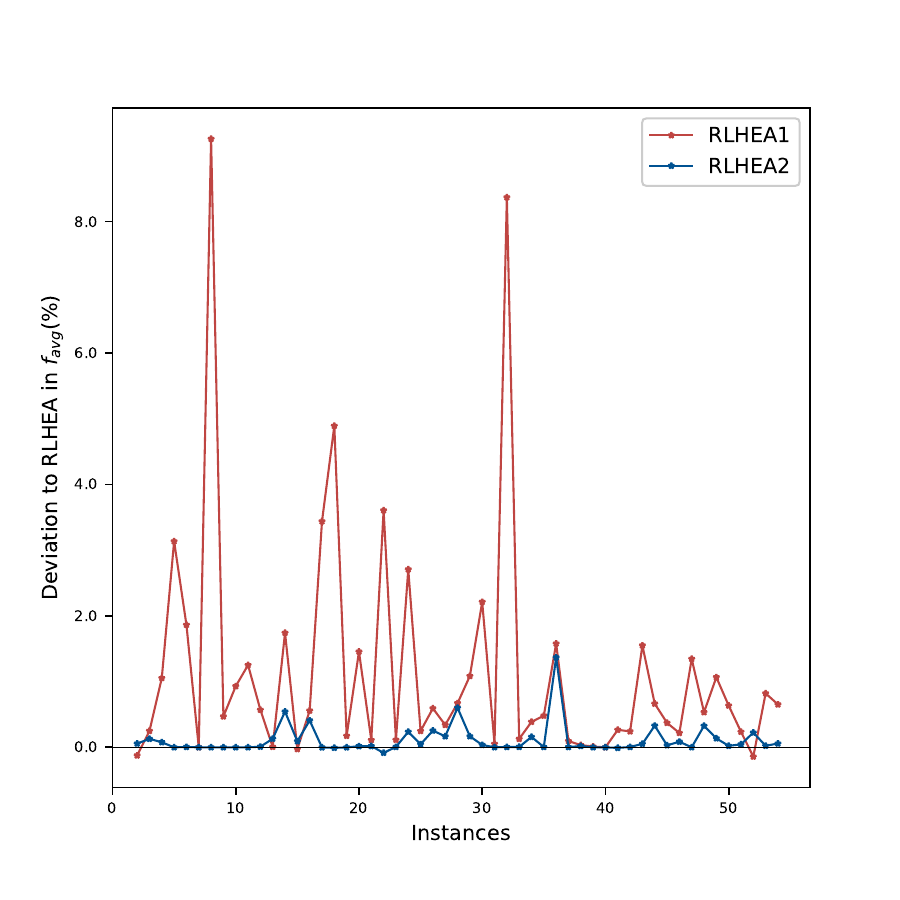}}
\end{minipage}
\caption{Comparative results of RLHEA with its two variants RLHEA1 and RLHEA2 on the 54 large instances.}
\label{fig mpeax compa}
\end{figure}

\begin{figure}[htbp]
\centering

\begin{minipage}[b]{0.49\textwidth}
  \centering
  \subfloat[1211122]{\label{fig conver1} \includegraphics[width=\textwidth]{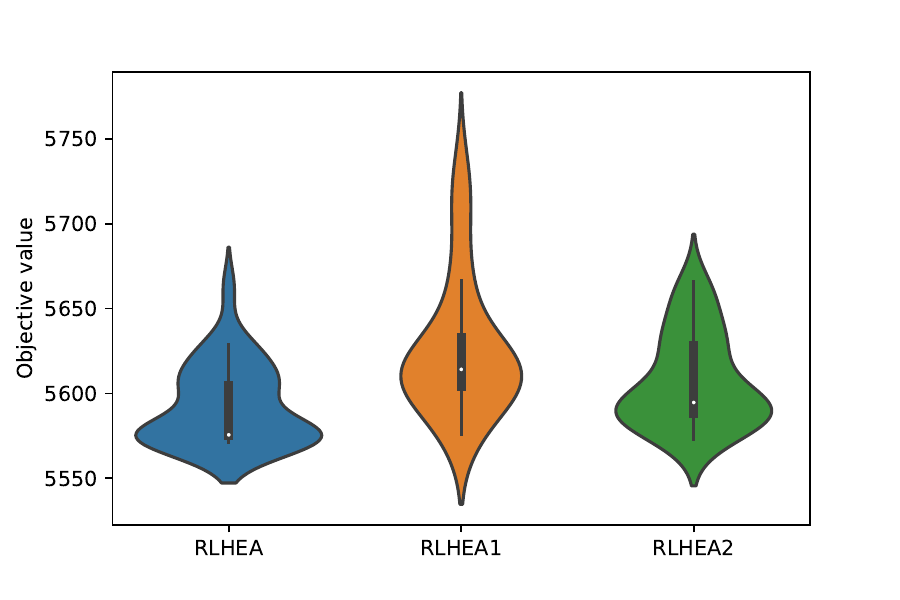}}
\end{minipage}
\begin{minipage}[b]{0.49\textwidth}
  \centering
  \subfloat[121212]{\label{fig conver2} \includegraphics[width=\textwidth]{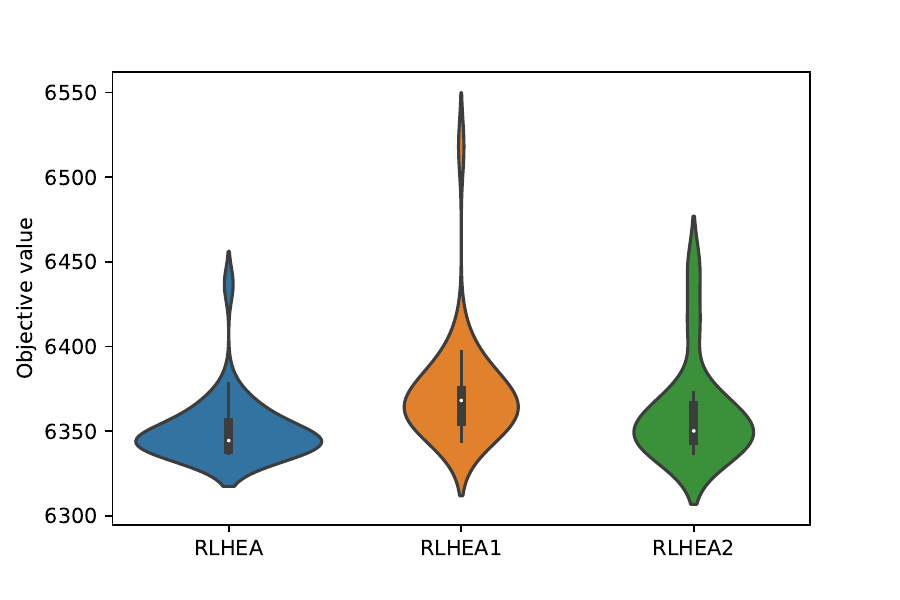}}
\end{minipage}

\begin{minipage}[b]{0.49\textwidth}
  \centering
  \subfloat[123122]{\label{fig conver3} \includegraphics[width=\textwidth]{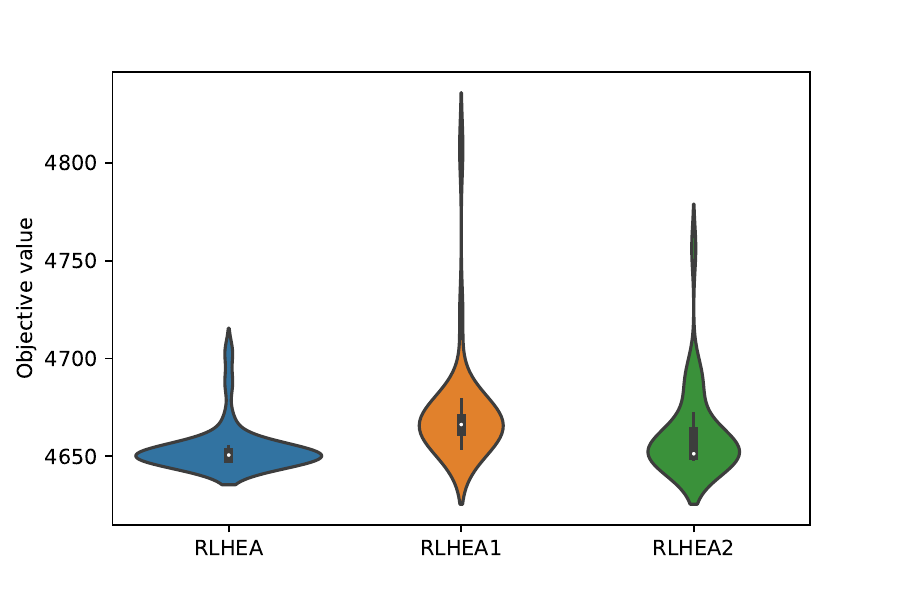}}
\end{minipage}
\begin{minipage}[b]{0.49\textwidth}
  \centering
  \subfloat[200-10-2]{\label{fig conver4}  \includegraphics[width=\textwidth]{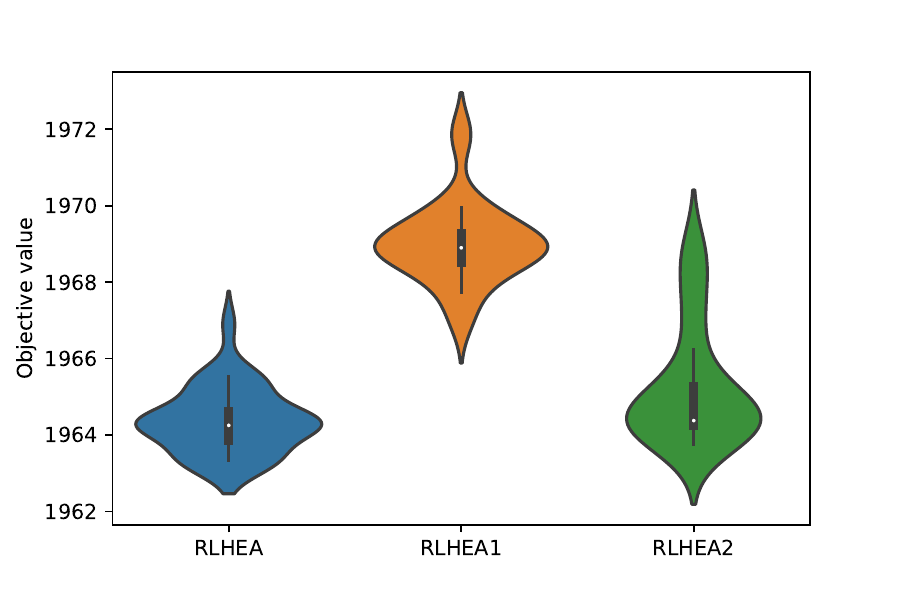}}
\end{minipage}

\caption{Violin plots for four instances of RLHEA, RLHEA1 and RLHEA2.}
\label{fig mpeax violin}
\end{figure}

Table \ref{tab benefit mpeax} shows the comparative results of RLHEA with the two variants in terms of the best and average objective values over 30 independent runs, along with the \textit{p}-values from the Wilcoxon signed-rank test. Fig. \ref{fig mpeax compa} shows the deviation of the two variants from the reference values given by RLHEA. In Fig. \ref{fig mpeax violin}, we present violin plots of the three algorithms on four large instances (121122, 121212, 123122, and 200-10-2), illustrating the distribution of objective values for the solutions obtained over the 30 independent runs. Looking at the results, it is evident that the RLHEA algorithm with the three-parent MPEAX crossover significantly outperforms the two variants, especially in terms of the average values, confirmed by the small \textit{p}-values. The violin plots clearly show that the solutions found by our RLHEA are more stable, demonstrating its robustness compared to the two variants. We conclude that RLHEA benefits from the edge assembly crossover method and the use of multiple parents in the crossover procedure.

\begin{table*}[htbp]
\renewcommand\tabcolsep{3pt}
\tiny
\begin{center}
\centering
\caption{Summarized comparison results of RLHEA against the RLHEA1 and RLHEA2 variants in terms of the best and average objective values on the 54 large instances.}
\label{tab benefit mpeax}

\begin{tabular}{llrrrrrrrr}
\toprule
\multirow{2}{*}{Instance}  &\multirow{2}{*}{ Pair algorithms}  &\multicolumn{4}{c}{$f_{best}$}                    &\multicolumn{4}{c}{$f_{avg}$}  \\
                                                               \cmidrule(r){3-6}                                   \cmidrule(r){7-10}
                            &                     & \#Wins   &\#Ties   &\#Losses   & \textit{p}-value             & \#Wins   &\#Ties   &\#Losses   & \textit{p}-value \\
\midrule
\multirow{2}{*}{Tuzun-Burke (36)}  &RLHEA vs. RLHEA1       &12         &24         & 0        &2.22e-3                          &33         & 1       & 2       &2.77e-2  \\
                                  &RLHEA vs. RLHEA2       &6          & 30        & 0        &4.95e-6                         &27         & 6        & 3        &7.68e-6  \\

\hline
\multirow{2}{*}{Prodhon (18)}     &RLHEA vs. RLHEA1       &7         &11        & 0        &1.80e-2                           &16        & 1       & 1       &6.00e-4  \\
                                 &RLHEA vs. RLHEA2       &2         &16        & 0       &0.18                               &15         &2       & 1        &9.35e-4 \\

\bottomrule
\end{tabular}
\end{center}
\end{table*}

\subsection{Benefits of Q-learning}
\label{subsec benefits q-learning}

As shown in Section \ref{subsubsec_sovnd}, the local search procedure uses Q-learning to determine the order in which the seven neighborhoods are explored. To evaluate the usefulness of this method, we created two algorithm variants, RLHEA3 and RLHEA4. The only difference between the two variants is the order of neighborhood exploration during local search, while the rest of the procedures remains the same. RLHEA3 uses a random order for neighborhood exploration, while RLHEA4 explores the neighborhoods in the fixed order $N_1$-$N_2$-$N_3$-$N_4$-$N_5$-$N_6$-$N_7$, which reflects the increasing complexity of these neighborhoods. We ran both algorithms on the 54 large instances as in Section \ref{subsec benefits mpeax}.

\begin{figure}[htbp]
\centering

\begin{minipage}[b]{0.49\textwidth}
  \centering
  \subfloat[131122]{\label{fig voilin6} \includegraphics[width=\textwidth]{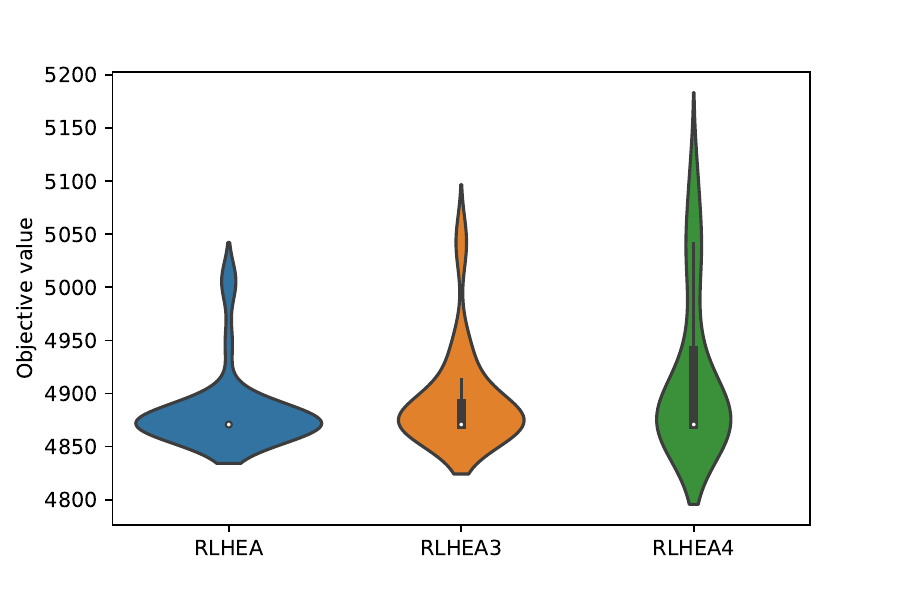}}
\end{minipage}
\begin{minipage}[b]{0.49\textwidth}
  \centering
  \subfloat[133122]{\label{fig voilin5} \includegraphics[width=\textwidth]{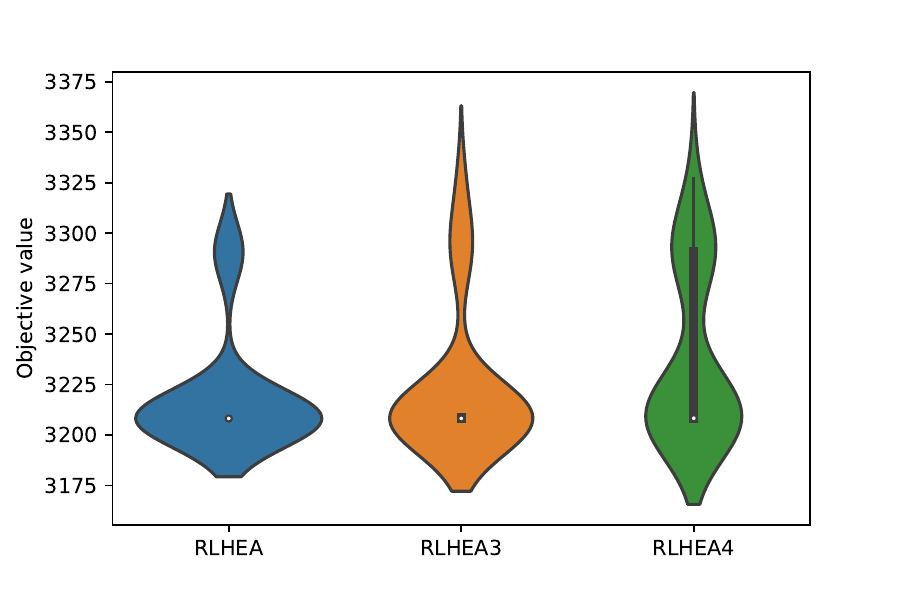}}
\end{minipage}

\begin{minipage}[b]{0.49\textwidth}
  \centering
  \subfloat[123122]{\label{fig voilin7} \includegraphics[width=\textwidth]{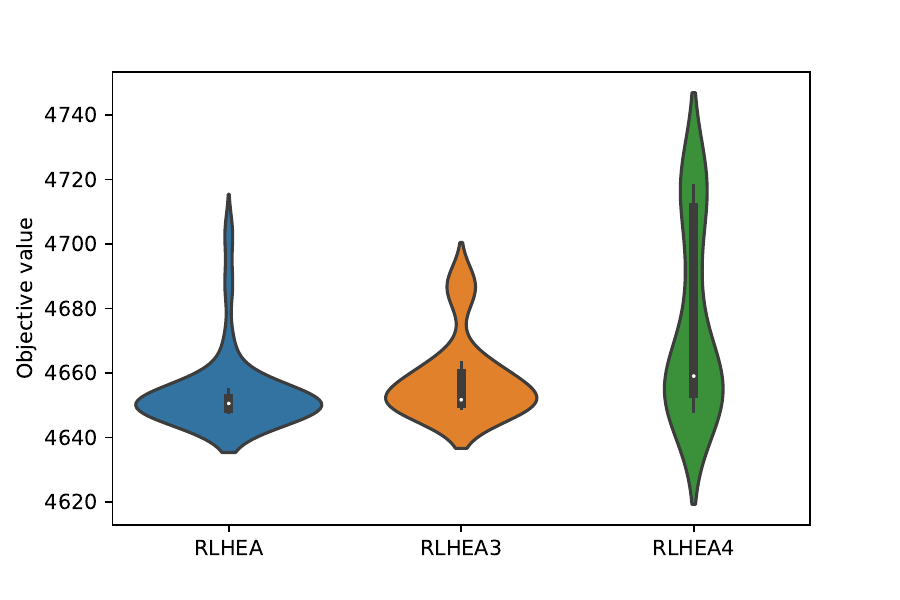}}
\end{minipage}
\begin{minipage}[b]{0.49\textwidth}
  \centering
  \subfloat[200-10-1b]{\label{fig voilin8}  \includegraphics[width=\textwidth]{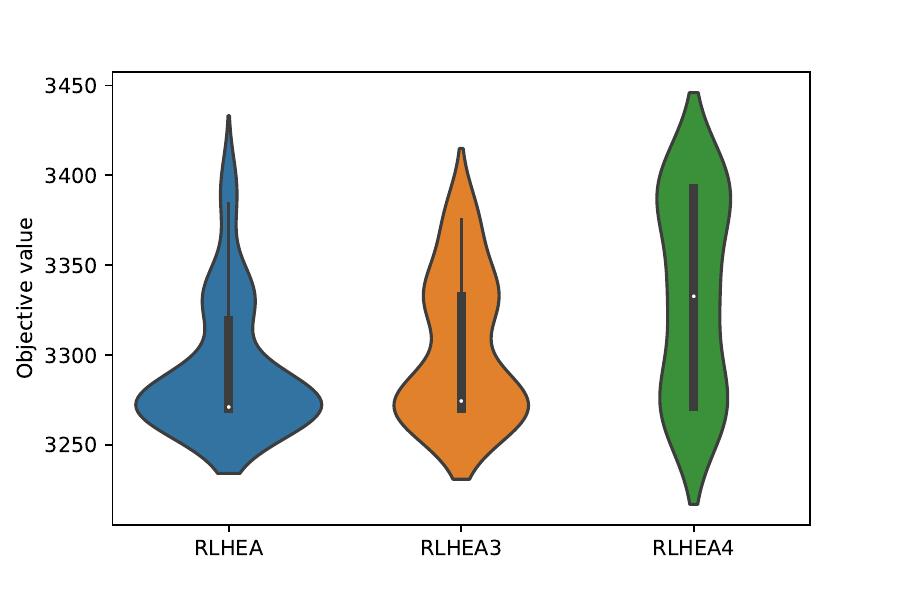}}
\end{minipage}

\caption{Violin plots for four instances of RLHEA, RLHEA3 and RLHEA4.}
\label{fig q-learning compa}
\end{figure}

Table \ref{tab benefit q-learning} summarizes the comparison of the results of RLHEA, RLHEA3 and RLHEA4 in terms of the best and the average objective values. Additionally, Fig. \ref{fig q-learning compa} shows violin plots for the three algorithms on four instances (131122, 133122, 123122, and 200-10-1b), indicating the deviation of the two variants from the reference values of RLHEA. The results show that RLHEA outperforms RLHEA3 and RLHEA4 in terms of the best objective values and especially in terms of the average objective values. The violin plots further show that the solution distribution obtained by RLHEA is more stable. In conclusion, the RLHEA algorithm performs better than the variants, benefiting from the Q-learning method to determine the exploration order of neighborhoods. %This clearly demonstrates the usefulness of Q-learning methods.

\begin{table*}[htbp]
\renewcommand\tabcolsep{3pt}
\tiny
\begin{center}
\centering
\caption{Summarized comparison results of RLHEA against the RLHEA3 and RLHEA4 variants in terms of the best and average objective values on the 54 large instances.}
\label{tab benefit q-learning}

\begin{tabular}{llrrrrrrrr}
\toprule
\multirow{2}{*}{Instance}  &\multirow{2}{*}{ Pair algorithms}  &\multicolumn{4}{c}{$f_{best}$}                    &\multicolumn{4}{c}{$f_{avg}$}  \\
                                                               \cmidrule(r){3-6}                                   \cmidrule(r){7-10}
                            &                     & \#Wins   &\#Ties   &\#Losses   & \textit{p}-value             & \#Wins   &\#Ties   &\#Losses   & \textit{p}-value \\
\midrule
\multirow{2}{*}{Tuzun-Burke(36)}  &RLHEA vs. RLHEA3       &3       &33         & 0       &0.11                            &28         & 4       & 4       &1.63e-5  \\
                                  &RLHEA vs. RLHEA4       &5       & 31        & 0       &0.04                            &36         & 0        & 0        &2.56e-6  \\

\hline
\multirow{2}{*}{Prodhon(18)}  &RLHEA vs. RLHEA3       &3         & 15        & 0        &0.11                         &12        & 3      &3       &7.55e-3  \\
                              &RLHEA vs. RLHEA4       &3         &15        & 0       &0.11                           &13         &3        & 2        &1.33e-3 \\

\bottomrule
\end{tabular}
\end{center}
\end{table*}

\begin{figure}[htbp]
\centering
\begin{minipage}[b]{0.49\textwidth}
  \centering
  \subfloat[Best objective value]{\label{fig influence mpeax subfig1}  \includegraphics[width=\textwidth]{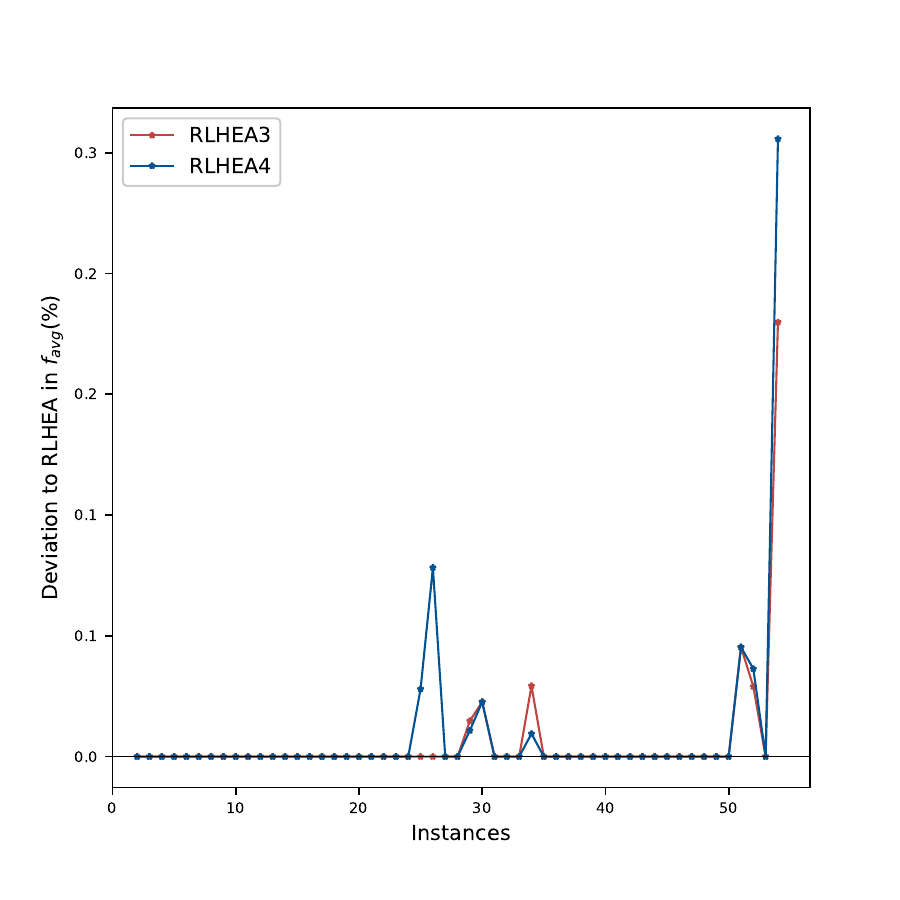}}
\end{minipage}
\hspace{0.000cm}
\begin{minipage}[b]{0.49\textwidth}
  \centering
  \subfloat[Average objective value]{\label{fig influence mpeax subfig2}  \includegraphics[width=\textwidth]{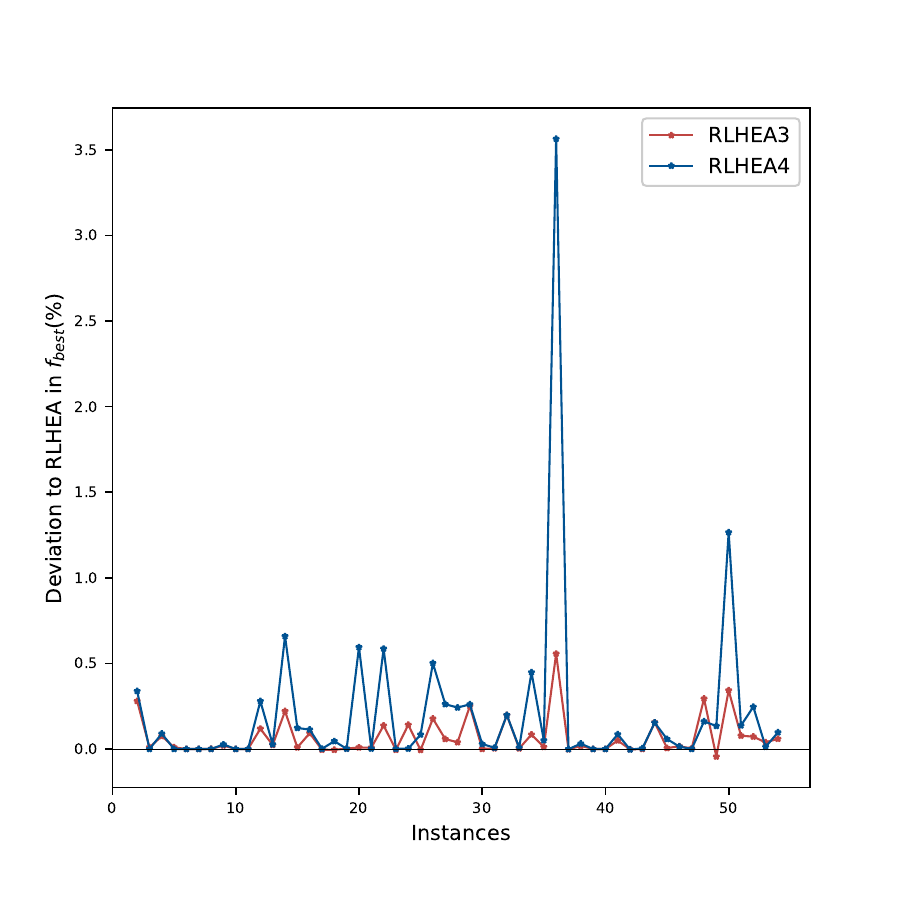}}
\end{minipage}
\caption{Comparative results of RLHEA with its two variants RLHEA3 and RLHEA4 on the 54 large instances.}
\label{fig q-learning violin}
\end{figure}

\subsection{Benefits of strategic oscillation}
\label{subsec benefits so}

As shown in Section \ref{subsec rlsovnd}, RLHEA uses strategic oscillation to examine both feasible and infeasible solutions by adaptively adjusting the penalty parameter $\beta$. To evaluate the benefit of this method, we created two variants, RLHEA5 and RLHEA6. RLHEA5 visits only feasible solutions, while RLHEA6 uses a fixed penalty parameter, which is set to the average cost per demand unit for each route of the VND input solution after the repair procedure. The summarized results of RLHEA, RLHEA5 and RLHEA6 are shown in Table \ref{tab benefit so}. From the results, we can see that compared to the two variants, RLHEA, which uses strategic oscillation to balance the visit of feasible and infeasible solutions, achieves better results in terms of the best and average objective values, showing the usefulness of the strategic oscillation method.

\begin{table*}[htbp]
\renewcommand\tabcolsep{3pt}
\tiny
\begin{center}
\centering
%\caption{Summary of the comparative results of the best and average objective values between RLHEA and the two algorithm variants RLHEA5, RLHEA6 on the 54 selected instances.}
\caption{Summarized comparison results of RLHEA against the RLHEA5 and RLHEA6 variants in terms of the best and average objective values on the 54 selected instances.}
\label{tab benefit so}
\begin{tabular}{llrrrrrrrr}
\toprule
\multirow{2}{*}{Instance}  &\multirow{2}{*}{ Pair algorithms}  &\multicolumn{4}{c}{$f_{best}$}                    &\multicolumn{4}{c}{$f_{avg}$}  \\
                                                               \cmidrule(r){3-6}                                   \cmidrule(r){7-10}
                            &                     & \#Wins   &\#Ties   &\#Losses   & \textit{p}-value             & \#Wins   &\#Ties   &\#Losses   & \textit{p}-value \\
\midrule
\multirow{2}{*}{Tuzun-Burke(36)}  &RLHEA vs. RLHEA5       &6       &30         & 0       &0.03                            &30         & 3      & 3       &5.91e-6  \\
                                  &RLHEA vs. RLHEA6       &3       &33        & 0       &0.11                            &26         & 6        &4        &1.24e-5  \\

\hline
\multirow{2}{*}{Prodhon(18)}  &RLHEA vs. RLHEA5       &3         & 15        & 0        &0.11                         &18        & 0      &0       &7.63e-6  \\
                              &RLHEA vs. RLHEA6       &2         &16        & 0       &0.18                           &14         &2        & 2        &1.12e-3 \\

\bottomrule
\end{tabular}
\end{center}
\end{table*}

\section{Conclusion}
\label{sec conclusion}

The latency location routing problem is a relevant model for various real-world problems, and a number of studies have proposed methods to solve this NP-hard problem. In this study, we introduced a reinforcement learning guided hybrid evolutionary algorithm to tackle this challenging problem. The algorithm consists of three key features. Its multi-parent edge assembly crossover with three parent solutions is capable of generating promising offspring solutions while enhancing solution diversity to mitigate diversity degradation by taking route orientation into account during the crossover procedure. Its Q-learning driven variable neighborhood descent dynamically determines the exploration order of multiple neighborhoods based on knowledge learned from the search history. The use of strategic oscillation during local optimization helps to dynamically visit different feasible search spaces by traversing infeasible search spaces.

We evaluated the proposed algorithm on the 76 benchmark instances commonly used in the literature and compared it with leading algorithms. Our approach achieved 51 new best solutions while matching the best known results for the remaining instances. In addition, we performed experiments to shed light on the key components of our algorithm and reveal the rationale behind these components.

The design principles behind the proposed algorithm are general and can be used to design effective algorithms for other problems. In particular, the idea of multi-parent edge assembly crossover is of interest for multi-route problems. The Q-learning technique used to determine the order of neighborhood exploration can contribute to the performance of local search methods that involve a portfolio of neighborhoods or search operators. Finally, the algorithm and its codes, which we will make publicly available, can be applied to practical applications related to the latency location routing problem.

\section*{Acknowledgments}

 We would like to express our gratitude to Dr. Alan Osorio-Mora and Prof. Paolo Toth, and their co-authors, for generously sharing the source codes of their works \citep{osorio2023iterated,osorio2023effective}, and their assistance in running the codes and addressing our inquiries. This work was partially supported by the National Natural Science Foundation Program of China (Grant No. 72122006). %Support from CSC (Grant No. 202106050037) for the first author is acknowledged.

\bibliographystyle{plain}
\bibliography{LLRP}
%\bibliographystyle{plain} 

%\bibliographystyle{elsarticle-num}
%\bibliography{LLRP}

\appendix

\section{Detailed comparison results }
\label{sec detailed results}

This section shows the detailed comparison results by the RLHEA algorithm and the reference algorithms, including GBILS \citep{nucamendi2022new}, SA-VND0, SA-VND1, SA-VND2 \citep{osorio2023effective} and M-ILS \citep{osorio2023iterated}. Tables \ref{table results tuzun} to \ref{table results barreto} show the comprehensive results on the three datasets, \textit{Tuzun-Burke}, \textit{Prodhon}, and \textit{Barreto}, respectively. The "Instances" column presents information about each instance, including the name, number of customers $N_c$, number of depots $N_d$ and the fleet size $N_v$.  It is worth noting that in the \textit{Prodhon} and \textit{Barreto} sets, the size information, such as the number of customers and depots, can be inferred from their names, so this information is not separately listed. The "BKS" column shows the current best-known objective value for each instance reported in the literature, where underlined values are proven optimal values. The columns $f_{best}$ and $f_{avg}$ indicate the best objective value and average value over all independent runs. $T_{avg}$ shows the average running time in seconds. It is worth noting that the presented running time for the algorithms SA-VND0, SA-VND1, SA-VND2, and M-ILS corresponds to the time on the same computer used for our RLHEA algorithm. To ensure a fair comparison, a scaling factor of 1.02 was applied to the running time of GBILS, based on their single-thread performance, as indicated by their machine information\footnote{https://www.cpubenchmark.net/}. The best solutions among the compared results are highlighted in bold in the tables. The improved best solutions (new upper bounds) are marked with an asterisk *.  
%\begin{sidewaystable}
\begin{landscape}
\tiny
\begin{table*}[htbp]
\renewcommand\tabcolsep{3pt}
\renewcommand{\baselinestretch}{1}\small\normalsize
\tiny
\begin{center}
\centering
\caption{Comparative results of the RLHEA algorithm with the reference algorithms on the set \textit{Tuzun-Burke}.}
\label{table results tuzun}
\begin{tabular}{lrrrllllllllllllllll}
    \toprule
    \multicolumn{5}{c}{Instances}       &\multicolumn{3}{c}{SA-VND0}   &\multicolumn{3}{c}{SA-VND1}      &\multicolumn{3}{c}{SA-VND2}     &\multicolumn{3}{c}{M-ILS}     &\multicolumn{3}{c}{RLHEA}\\
    \cmidrule(r){1-5}                                              \cmidrule(r){6-8}                    \cmidrule(r){9-11}                   \cmidrule(r){12-14}         \cmidrule(r){15-17}                     \cmidrule(r){18-20}
Name &$N_c$ &$N_d$  &$N_v$   &BKS   &$f_{best}$ &$f_{avg}$ &$T_{avg}$    &$f_{best}$ &$f_{avg}$ &$T_{avg}$       &$f_{best}$ &$f_{avg}$ &$T_{avg}$        &$f_{bst}$ &$f_{avg}$ &$T_{avg}$    &$f_{best}$ &$f_{avg}$ &$T_{avg}$    \\
    \midrule
        111112 &100 &10 &11 & 3834.91   & 3862.86 & 3971.50 & 164.54    & 3892.97 & 3972.93 & 213.53     & 3887.86 & 3970.09 & 234.55    & 3834.91 & 3884.56 & 172.92     &\bf3826.78* & 3827.15 & 81.67 \\ 
        111122 &100 &20 &11 & 3602.70   & 3612.36 & 3694.70 & 168.76    & 3633.60 & 3712.64  & 218.58    & 3602.70 & 3693.00 & 252.69    & 3659.46 & 3698.24 & 182.33     &\bf3597.64* & 3604.37 & 78.18 \\ 
        111212 &100 &10 &10 & 3919.74   & 3960.24 & 4038.24 & 154.78    & 3988.11 & 4067.91 & 208.51     & 3963.34 & 4054.02 & 222.67    & 3919.74 & 3998.66 & 203.02     &\bf3901.18* & 3902.75 & 101.06 \\ 
        111222 &100 &20 &11 & 4065.04   & 4086.74 & 4140.33 & 170.05    & 4077.87 & 4147.90 & 211.86     & 4065.70 & 4135.25 & 249.10    & 4065.04 & 4134.88 & 182.71     &\bf4058.09* & 4062.18 & 86.99 \\ 
        112112 &100 &10 &11 & 2726.41   & 2739.16 & 2755.53 & 206.47    & 2740.21 & 2759.43 & 263.28     & 2749.48 & 2757.67 & 286.65    &\bf2726.41 & 2749.93 & 143.99   &\bf2726.41  & 2726.41 & 85.78 \\ 
        112122 &100 &20 &11 & 2057.30   & 2060.29 & 2072.57 & 198.98    & 2057.45 & 2078.03 & 243.93     & 2060.29 & 2074.93 & 273.22    & 2057.30 & 2061.79 & 96.96      &\bf2056.84* &2056.84  &60.14  \\ 
        112212 &100 &10 &12 & 1394.65   & 1402.97 & 1416.20 & 208.88    & 1403.57 & 1416.72 & 280.59     & 1404.58 & 1415.56 & 300.11    &\bf1394.65 & 1409.73 & 136.47   &\bf1394.65  & 1394.65 & 56.27 \\ 
        112222 &100 &20 &11 & 1618.93   & 1623.69 & 1633.00 & 219.86    & 1621.40 & 1633.94 & 276.28     & 1625.44 & 1634.90 & 293.26    & 1618.93 & 1631.37 & 139.87     &\bf1614.83* & 1614.83 & 78.51 \\ 
        113112 &100 &10 &11 & 2826.52   & 2837.51 & 2852.63 & 182.80    & 2835.76 & 2853.57 & 241.31     & 2833.66 & 2853.50 & 268.61    &\bf2826.52 & 2841.15 & 158.13   &\bf2826.52  & 2826.52 & 104.47 \\ 
        113122 &100 &20 &11 & 2772.98   & 2776.39 & 2782.53 & 175.52    & 2774.36 & 2784.42 & 246.00     & 2776.39 & 2781.45 & 251.11    &\bf2772.98 & 2798.60 & 214.48   &\bf2772.98  & 2772.98 & 65.16 \\ 
        113212 &100 &10 &12 & 1815.62   & 1817.00 & 1823.15 & 189.94    &\bf 1815.62 & 1822.81 & 256.71  &\bf 1815.62 & 1823.88 & 277.66 & 1817.00 & 1835.52 & 181.58     &\bf1815.62  & 1815.62 & 47.89 \\ 
        113222 &100 &20 &11 & 1876.14   & 1876.14 & 1888.46 & 170.52    & 1879.63 & 1890.96 & 217.34     & 1878.17 & 1891.00 & 233.39    & 1876.58 & 1885.67 & 146.63     &\bf1874.42* & 1874.45 & 68.21 \\ 
        131112 &150 &10 &16 & 5411.43   & 5473.18 & 5582.94 & 320.34    & 5464.21 & 5570.12 & 507.95     & 5466.75 & 5589.24 & 545.67    & 5411.43 & 5478.14 & 319.87     &\bf5405.04* & 5406.30 & 174.53 \\ 
        131122 &150 &20 &16 & 4926.87   & 4993.36 & 5142.06 & 364.16    & 5009.26 & 5143.19 & 541.17     & 4967.39 & 5154.30 & 547.68    & 4926.87 & 5051.01 & 314.02     &\bf4870.82* &4884.55 &182.04 \\ 
        131212 &150 &10 &17 & 5558.83   & 5679.70 & 5787.34 & 371.24    & 5606.31 & 5785.18 & 603.00     & 5658.71 & 5776.08 & 564.77    & 5558.83 & 5637.11 & 341.80     &\bf5525.91* & 5550.53 & 156.06 \\ 
        131222 &150 &20 &17 & 5060.71   & 5141.89 & 5284.44 & 366.60    & 5126.95 & 5277.65 & 580.13     & 5166.72 & 5279.36 & 529.19    & 5060.71 & 5106.51 & 350.46     &\bf5039.22* & 5068.44 & 180.93 \\ 
        132112 &150 &10 &16 & 3850.90   & 3868.88 & 3895.91 & 491.93    & 3883.40 & 3899.41 & 757.98     & 3879.81 & 3901.40 & 728.69    & 3850.90 & 3881.92 & 305.86     &\bf3831.89* & 3832.05 & 218.87 \\ 
        132122 &150 &20 &16 & 3738.61   & 3740.10 & 3795.93 & 431.31    & 3752.76 & 3787.72 & 654.21     & 3768.60 & 3796.56 & 667.73    & 3738.61 & 3785.14 & 304.60     &\bf3721.93* & 3722.67 & 176.88 \\ 
        132212 &150 &10 &17 & 2835.66   & 2842.10 & 2857.29 & 519.25    & 2837.84 & 2860.70 & 741.29     & 2840.11 & 2855.58 & 761.57    & 2835.66 & 2848.89 & 275.57     &\bf2835.25* & 2835.34 & 139.13 \\ 
        132222 &150 &20 &17 & 1655.39   & 1660.89 & 1691.97 & 561.19    & 1672.86 & 1697.45 & 761.57     & 1669.51 & 1691.96 & 778.57    & 1655.39 & 1676.46 & 215.95     &\bf1646.45* & 1646.56 & 178.13 \\ 
        133112 &150 &10 &16 & 4581.60   & 4588.38 & 4619.91 & 399.98    & 4598.23 & 4630.69 & 572.31     & 4587.35 & 4633.07 & 596.78    & 4581.60 & 4615.95 & 235.00     &\bf4556.33* & 4556.33 & 149.38 \\ 
        133122 &150 &20 &16 & 3211.98   & 3223.44 & 3259.45 & 438.16    & 3225.56 & 3271.04 & 677.99     & 3227.27 & 3255.20 & 630.51    & 3211.98 & 3237.28 & 310.77     &\bf3208.21* & 3208.60 & 178.17 \\ 
        133212 &150 &10 &17 & 2903.36   & 2911.58 & 2938.05 & 531.82    & 2911.35 & 2938.01 & 761.78     & 2905.45 & 2938.41 & 769.77    & 2903.36 & 2919.40 & 246.40     &\bf2896.63* & 2897.15 & 151.80 \\ 
        133222 &150 &20 &17 & 2485.07   & 2502.97 & 2551.31 & 501.77    & 2502.68 & 2558.34 & 728.81     & 2507.87 & 2534.09 & 724.39    & 2485.07 & 2492.94 & 231.98     &\bf2484.68* & 2484.68 & 182.02 \\ 
        121112 &200 &10 &21 & 6573.72   & 6608.45 & 6821.20 & 762.24    & 6621.55 & 6881.38 & 1196.02    & 6721.82 & 6894.98 & 1180.02   & 6573.72 & 6628.07 & 502.90     &\bf6499.53* & 6506.93 & 316.29 \\ 
        121122 &200 &20 &22 & 5612.82   & 5730.53 & 5954.23 & 892.22    & 5788.72 & 5966.38 & 1392.03    & 5762.73 & 5935.57 & 1479.41   & 5612.82 & 5679.77 & 487.83     &\bf5571.24* & 5592.63 & 320.55 \\ 
        121212 &200 &10 &21 & 6394.33   & 6503.36 & 6613.84 & 791.44    & 6429.62 & 6608.92 & 1258.20    & 6512.62 & 6643.75 & 1237.28   & 6394.33 & 6450.87 & 505.81     &\bf6337.03* & 6350.82 & 312.35 \\ 
        121222 &200 &20 &21 & 6428.31   & 6551.73 & 6759.22 & 789.09    & 6562.11 & 6796.71 & 1324.77    & 6544.89 & 6769.32 & 1242.46   & 6428.31 & 6522.74 & 573.24     &\bf6349.27* & 6382.93 & 320.36 \\ 
        122112 &200 &10 &21 & 6111.52   & 6154.64 & 6255.10 & 883.92    & 6184.70 & 6280.31 & 1705.19    & 6192.70 & 6274.04 & 1153.85   & 6111.52 & 6203.24 & 833.95     &\bf6018.59* &6043.38 &486.10  \\ 
        122122 &200 &20 &21 & 3726.80   & 3757.37 & 3782.47 & 995.03    & 3757.27 & 3792.67 & 1547.20    & 3751.31 & 3786.07 & 1459.85   & 3726.80 & 3754.67 & 437.73     &\bf3705.59* & 3709.02 & 264.10 \\ 
        122212 &200 &10 &21 & 4018.86   & 4046.81 & 4075.78 & 922.81    & 4046.42 & 4078.60 & 1520.92    & 4048.88 & 4082.98 & 1352.78   & 4018.86 & 4036.49 & 380.40     &\bf4013.55* & 4014.10 & 265.99 \\ 
        122222 &200 &20 &22 & 2047.95   & 2054.32 & 2083.56 & 996.79    & 2052.22 & 2084.10 & 1530.59    & 2061.32 & 2081.97 & 1490.62   & 2047.95 & 2057.11 & 358.11     &\bf2033.34* & 2033.52 & 316.88 \\ 
        123112 &200 &10 &22 & 4868.90   & 4916.97 & 5024.07 & 908.81    & 4967.11 & 5047.87 & 1407.80    & 4931.24 & 5019.94 & 1489.04   & 4868.90 & 4916.56 & 520.88     &\bf4842.05* & 4842.08 & 313.68 \\ 
        123122 &200 &20 &22 & 4675.34   & 4725.91 & 4771.90 & 898.15    & 4707.60 & 4785.89 & 1350.76    & 4719.04 & 4761.84 & 1278.30   & 4675.34 & 4705.53 & 545.92     &\bf4647.68* & 4654.10 & 320.55 \\ 
        123212 &200 &10 &22 & 5135.21   & 5170.77 & 5218.43 & 868.24    & 5178.02 & 5225.04 & 1351.98    & 5183.34 & 5259.76 & 1172.57   & 5135.21 & 5174.88 & 332.78     &\bf5123.41* & 5124.13 & 262.12 \\ 
        123222 &200 &20 &22 & 2528.74   & 2567.20 & 2629.66 & 876.49    & 2555.18 & 2633.86 & 1363.16    & 2562.46 & 2602.87 & 1353.88   & 2528.74 & 2557.17 & 344.07     &\bf2494.58* & 2494.58 & 263.87 \\ 
   \bottomrule
\end{tabular}
\end{center}
\end{table*}
\end{landscape}

\begin{landscape}
%\begin{sidewaystable}[ht]
\begin{table*}[htbp]
\renewcommand\tabcolsep{3pt}
\renewcommand{\baselinestretch}{1}\small\normalsize
\tiny

\centering
\caption{Comparative results of proposed RLHEA with the reference algorithms on the set \textit{Prodhon}.}
\label{table results prodhon}

\begin{tabular}{lrr lllllllllllllllll}
    \toprule
    \multicolumn{3}{c}{Instances}   &\multicolumn{2}{c}{GBILS}    &\multicolumn{3}{c}{SA-VND0}   &\multicolumn{3}{c}{SA-VND1}      &\multicolumn{3}{c}{SA-VND2}     &\multicolumn{3}{c}{M-ILS}     &\multicolumn{3}{c}{RLHEA}\\
    \cmidrule(r){1-3}                \cmidrule(r){4-5}             \cmidrule(r){6-8}                \cmidrule(r){9-11}         \cmidrule(r){12-14}                     \cmidrule(r){15-17}         \cmidrule(r){18-20}
Name &$N_v$   &BKS     &$f_{best}$  &$T_{avg}$    &$f_{best}$ &$f_{avg}$ &$T_{avg}$    &$f_{best}$ &$f_{avg}$ &$T_{avg}$       &$f_{best}$ &$f_{avg}$ &$T_{avg}$        &$f_{bst}$ &$f_{avg}$ &$T_{avg}$    &$f_{best}$ &$f_{avg}$ &$T_{avg}$    \\
    \midrule
        20-5-1    &5   & \underline{330.00}  &\bf330.00 & 0.34   &\bf330.00 & 330.00 & 9.85      &\bf330.00 & 330.00 & 8.59       &\bf330.00 & 330.00 & 13.11      &\bf330.00 & 330.00 & 13.92     &\bf330.00 & 330.00 & 3.23 \\ 
        20-5-1b   &3   & \underline{608.05} &\bf608.06 & 0.15    &\bf608.05 & 608.06 & 11.56     &\bf608.05 & 608.06 & 8.80       & 608.05 & 608.06 & 19.98        & 615.66 & 615.66   & 10.10     &\bf608.05 & 608.06 & 3.57 \\ 
        20-5-2    &5   & \underline{301.97} &\bf301.97 &0.24     &\bf301.97 & 301.97 & 8.31      &\bf301.97 & 301.97 & 7.28       &\bf301.97 & 301.97 & 12.27      &\bf301.97 & 301.97 & 13.64     &\bf301.97 & 301.97 & 3.76 \\ 
        20-5-2b   &3   &\underline{ 486.55} &\bf486.55 &0.25     &\bf486.55 & 486.55 & 12.31     &\bf486.55 & 486.55 & 10.23      &\bf486.55 &486.55 & 20.68       &\bf486.55 & 486.55 & 10.10     &\bf486.55 & 486.54 & 3.69 \\ 
        50-5-1    &12  & \underline{843.93} & 846.88  &50.32     & 846.17   & 849.77 & 59.10     & 846.51 & 850.10 & 58.17        & 844.63   & 849.66 & 69.99      &\bf843.93 & 845.75 & 48.57     &\bf843.93 & 843.93 & 15.71 \\ 
        50-5-1b   &6   & \underline{1293.46} &\bf1293.93 &21.46  &\bf1293.46 & 1293.71 & 46.72   &\bf1293.46 & 1293.54 & 44.53    &\bf1293.46 & 1293.48 & 73.74    &\bf1293.46 & 1293.95 & 58.81   &\bf1293.46 & 1293.46 & 20.90 \\ 
        50-5-2    &12  &\underline{ 684.13} & 691.67 &38.49      &\bf684.13 & 693.78 & 48.30     &\bf684.13 & 692.43 & 60.38      &\bf684.13 & 697.61 & 59.33      &\bf684.13 & 690.69 & 73.43     &\bf684.13 & 684.13 & 15.72 \\ 
        50-5-2b   &6   &\underline{ 953.25} & 954.88 &22.44      &\bf953.25 & 953.50 & 39.05     &\bf953.25 & 953.35 & 39.71      &\bf953.25 & 953.40 & 63.61      &\bf953.25 & 953.68 & 48.90     &\bf953.25 & 953.25 & 16.73 \\ 
        50-5-2BIS &12  & \underline{945.45} & 952.55 &39.56      & 949.13 & 950.80 & 65.24       & 949.57 & 951.13 & 83.25        & 949.56 & 951.46 & 73.98        &\bf945.45 & 945.65 & 121.85    &\bf945.45 & 945.45 & 12.14 \\ 
        50-5-2bBIS &6  & \underline{803.90} &\bf803.90 &31.67    &\bf803.90 & 803.90 & 49.70     &\bf803.90 & 803.90 & 48.13      &\bf803.90 & 803.90 & 68.76      &\bf803.90 & 803.90 & 54.99     &\bf803.90 & 803.90 & 13.73 \\ 
        50-5-3    &12  & 831.57 & 832.15 &39.09      &831.97 & 835.10 & 53.74     & 833.01 & 834.91 & 62.43        & 833.01 & 836.25 & 69.43        &\bf831.57 & 834.30 & 81.44     &\bf831.57 & 831.57 & 18.37 \\ 
        50-5-3b   &6   & \underline{1101.57} & 1106.57 &22.75    &\bf1101.57 & 1103.15 & 39.30   &\bf1101.57 & 1103.94 & 41.72    &\bf1101.57 & 1102.47 & 61.71    &\bf1101.57 & 1102.49 & 37.31   &\bf1101.57 & 1101.57 & 19.59 \\ 
        100-5-1   &24  & 2000.80 & 2035.60 &21.24    & 2004.33 & 2023.35 & 246.05    & 2010.49 & 2023.78 & 378.85     & 2016.44 & 2028.27 & 307.25     & 2000.80 & 2012.06 & 184.89    &\bf1997.29* & 1997.37 & 63.14 \\ 
        100-5-1b  &12  & 2311.01 & 2357.87 &33.69    & 2311.84 & 2336.64 & 182.67    & 2312.53 & 2337.27 & 227.13     & 2317.42 & 2336.82 & 252.54     & 2311.01 & 2346.47 & 171.41    &\bf2305.65* & 2305.89 & 68.43 \\ 
        100-5-2   &24  & 1128.12 & 1144.70 &26.60     & 1132.36 & 1135.99 & 214.42   & 1129.83 & 1135.49 & 343.26     & 1132.68 & 1136.61 & 282.63     & 1128.12 & 1133.17 & 198.13    &\bf1126.39* & 1126.39 & 65.69 \\ 
        100-5-2b  &11  & 1507.88 & 1567.44 &31.73     & 1507.88 & 1517.11 & 205.25   & 1510.57 & 1519.04 & 247.41     & 1510.24 & 1519.44 & 271.72     & 1507.88 & 1511.89 & 119.62    &\bf1506.79* & 1506.79 & 84.88 \\ 
        100-5-3   &24  & 1572.61 & 1596.77 &16.12     & 1581.93 & 1587.20 & 219.07   & 1581.93 & 1586.49 & 344.91     & 1579.38 & 1587.41 & 283.76     & 1572.61 & 1582.05 & 233.51    &\bf1567.62* & 1568.22 & 59.34 \\ 
        100-5-3b  &11  & 1933.70 & 2032.13 &37.66     & 1933.70 & 1950.89 & 180.90   & 1935.70 & 1953.85 & 258.79     & 1940.47 & 1955.56 & 255.64     & 1934.93 & 1954.50 & 163.38    &\bf1932.96* & 1933.07 & 73.12 \\ 
        100-10-1  &26  & 1458.80 & 1481.56 &26.33     & 1472.85 & 1511.00 & 247.27   & 1470.71 & 1503.92 & 371.15     & 1461.53 & 1513.44 & 333.90     & 1458.80 & 1464.80 & 196.75    &\bf1457.53* & 1457.68 & 59.80 \\ 
        100-10-1b &12  & 1899.80 & 1984.91 &33.05     & 1901.27 & 1953.96 & 193.46   & 1915.77 & 1972.26 & 253.50     & 1926.32 & 1963.33 & 255.59     & 1899.80 & 1918.20 & 186.81    &\bf1894.92* & 1895.83 & 63.40 \\ 
        100-10-2  &24  & 1137.59 & 1287.50 &24.84     & 1143.30 & 1152.81 & 251.82   & 1142.31 & 1155.47 & 370.05     & 1141.45 & 1156.81 & 332.11     &1137.59  & 1144.99 & 233.59    &\bf1134.80* & 1135.23 & 48.99 \\ 
        100-10-2b &11  & 1559.88 & 1645.07 &45.62    & 1566.48 & 1585.67 & 195.28    & 1566.48 & 1588.24 & 256.10     & 1568.71 & 1583.06 & 267.51     & 1559.88 & 1570.80 & 190.30    &\bf1555.71* & 1555.71 & 58.02 \\ 
        100-10-3  &25  & 1204.94 & 1216.20 &18.76     & 1209.20 & 1221.52 & 277.89   & 1209.86 & 1225.98 & 376.80     & 1210.61 & 1225.49 & 352.77     &1204.94  & 1209.27 & 151.09    &\bf1204.01* & 1204.01 & 56.63 \\ 
        100-10-3b &11  & 1653.83 & 1745.05 &22.99    & 1662.43 & 1705.63 & 187.74    & 1665.69 & 1706.68 & 243.04     & 1676.25 & 1705.51 & 262.94     & 1653.83 & 1670.95 & 196.73    &\bf1647.85* & 1649.38 & 67.45 \\ 
        200-10-1  &49  & 2780.03 & 2861.85 &91.20    & 2798.57 & 2854.10 & 1327.81   & 2797.86 & 2863.27 & 2204.55    & 2792.24 & 2860.29 & 2026.86    & 2780.03 & 2788.95 & 568.99    &\bf2770.45* & 2774.33 & 240.31 \\ 
        200-10-1b &22  & 3290.73 & 3557.96 &99.69    & 3368.71 & 3477.07 & 956.22    & 3355.70 & 3478.91 & 1572.79    & 3327.76 & 3456.43 & 1411.39    & 3290.73 & 3336.49 & 522.54    &\bf3270.68* & 3292.13 & 249.43 \\ 
        200-10-2  &49  & 1973.41 & 1997.01 &112.28   & 1984.96 & 2001.97 & 1164.73   & 1986.55 & 2004.51 & 2571.48    & 1986.51 & 2003.80 & 1853.75    & 1973.41 & 1980.50 & 429.89    &\bf1963.32* & 1964.37 & 206.21 \\ 
        200-10-2b &23  & 2328.12 & 2473.24 &89.74    & 2336.11 & 2379.01 & 932.35    & 2355.15 & 2378.21 & 1704.98    & 2336.29 & 2377.63 & 1453.02    & 2328.12 & 2360.59 & 438.38    &\bf2309.30* & 2314.74 & 174.98 \\ 
        200-10-3  &48  & 2727.15 & 2783.20 &106.18    & 2741.16 & 2758.09 & 1075.86  & 2744.67 & 2757.42 & 2293.23    & 2750.18 & 2762.07 & 1612.10    & 2727.15 & 2736.47 & 419.99    &\bf2719.34* & 2720.27 & 182.54 \\ 
        200-10-3b &22  & 3194.53 & 3413.34 &92.27    & 3242.18 & 3274.58 & 772.47    & 3233.89 & 3267.81 & 1422.02    & 3225.11 & 3272.52 & 1226.53    & 3194.53 & 3220.29 & 345.12    &\bf3174.91* & 3186.06 & 183.21 \\  
   \bottomrule
  
\end{tabular}
%\end{sidewaystable}
\end{table*}
%\end{landscape}

%\begin{landscape}
\begin{table*}[htbp]
\renewcommand\tabcolsep{3pt}
\renewcommand{\baselinestretch}{1}\small\normalsize
\tiny
\begin{center}
\centering
\caption{Comparative results of proposed RLHEA with the reference algorithms on the set \textit{Barreto}.}
\label{table results barreto}
\begin{tabular}{lrrlllllllllllllllllll}
    \toprule
    \multicolumn{3}{c}{Instances}   &\multicolumn{2}{c}{GBILS}    &\multicolumn{3}{c}{SA-VND0}   &\multicolumn{3}{c}{SA-VND1}      &\multicolumn{3}{c}{SA-VND2}     &\multicolumn{3}{c}{M-ILS}     &\multicolumn{3}{c}{RLHEA}\\
    \cmidrule(r){1-3}                \cmidrule(r){4-5}             \cmidrule(r){6-8}                \cmidrule(r){9-11}         \cmidrule(r){12-14}                     \cmidrule(r){15-17}         \cmidrule(r){18-20}
Name   &$N_v$   &BKS     &$f_{best}$  &$T_{avg}$    &$f_{best}$ &$f_{avg}$ &$T_{avg}$    &$f_{best}$ &$f_{avg}$ &$T_{avg}$       &$f_{best}$ &$f_{avg}$ &$T_{avg}$        &$f_{bst}$ &$f_{avg}$ &$T_{avg}$    &$f_{best}$ &$f_{avg}$ &$T_{avg}$    \\
    \midrule
        Christ\_50\_5   &6  & \underline{1661.64}   & 1719.89 & 11.99     &\bf1661.64 & 1662.06 & 46.25     &\bf1661.64 & 1662.13 & 42.96    &\bf1661.64 & 1662.13 & 64.7   &\bf1661.64 & 1669.05 & 51.88    &\bf1661.64 & 1661.64 & 18.65 \\ 
        Christ\_75\_10  &9 & 2370.73 & 2399.28 & 51.12        & 2403.79 & 2459.03 & 96.45       & 2383.04 & 2457.45 & 122.18     & 2400.86 & 2459.58 & 128.00   &2370.73 & 2417.28 & 127.97      &\bf2362.48* & 2362.48 & 40.98 \\ 
        Christ\_100\_10 &8 & 3791.98 & 3984.05 & 148.17       & 3791.98 & 3831.18 & 177.42      & 3806.39 & 3838.89 & 198.76     & 3797.00 & 3826.81 & 231.22   &3803.5 & 3848.63 & 168.75       &\bf3788.96* & 3788.96 & 92.92 \\ 
        Gaskell\_21\_5  &4 & \underline{653.48} &\bf653.48 & 0.28        &\bf653.48 & 653.48 & 9.33         &\bf653.48 & 653.48 & 8.67       &\bf653.48 & 653.48 & 13.86    &\bf653.48 &653.48 & 11.93       &\bf653.48 &653.48 & 4.06 \\ 
        Gaskell\_29\_5  &4 & \underline{1199.33} &\bf1199.33 & 1.70      &\bf1199.33 &1199.33 & 25.33       &\bf1199.33 &1199.33 & 21.29     &\bf1199.33 &1199.33 & 40.27   &\bf1199.33 &1199.33 & 22.26     &\bf1199.33 & 1199.33 & 7.30 \\ 
        Gaskell\_32\_5b &3 & \underline{1552.84} &\bf1552.84 & 1.60      &\bf1552.84 & 1553.29 & 35.58      &\bf1552.84 & 1553.29 & 27.65    &\bf1552.84 & 1552.84 & 55.69  &\bf1552.84 & 1556.58 & 24.35    &\bf1552.84 & 1552.84 & 9.00 \\ 
        Gaskell\_36\_5  &4 & \underline{1627.17} &\bf1627.17 & 1.05      &\bf1627.17 & 1627.17 & 25.76      &\bf1627.17 & 1627.17 & 22.72    &\bf1627.17 & 1627.17 & 39.59  &\bf1627.17 & 1628.12 & 29.56    &\bf1627.17 & 1627.17 & 11.26 \\ 
        Min\_27\_5      &4 & \underline{5387.55} &\bf5387.55 & 27.55     &\bf5387.55 & 5387.55 & 14.27      &\bf5387.55 & 5387.55 & 12.25    &\bf5387.55 & 5387.55 & 22.73  &\bf5387.55 & 5387.55 & 10.59    &\bf5387.55 & 5387.55 & 5.98 \\ 
        Min\_134\_8     &11 & 21751.97 &- & -        & 21852.35 & 22307.28 & 194.27     & 21910.54 & 22309.19 & 215.35    & 21853.52& 22256.20 & 276.15    &\bf21751.97 & 22450.39 & 164.69      &\bf21751.97 & 21751.97 & 131.91 \\ 
        Or\_117\_14     &7 & 53798.53 &- &- &\bf53798.53 & 54866.72 & 108.95       & 53859.08 & 54805.87 & 105.50     & 54103.16 & 54902.35 & 171.64     & 54328.75 & 56687.87 & 125.96          &\bf53798.53 & 53907.31 & 97.84 \\ 
   \bottomrule
\end{tabular}
\end{center}
\end{table*}
\end{landscape}

\end{document}